%% file: main.tex
\documentclass[11pt]{article}

\usepackage[final]{acl}

\usepackage{times}
\usepackage{latexsym}
\usepackage[T1]{fontenc}
\usepackage[utf8]{inputenc}
\usepackage{inconsolata}
\usepackage{graphicx}

\usepackage{amsmath}
\usepackage{amssymb}
\usepackage{amsfonts}
\usepackage{xcolor}

\usepackage{booktabs}
\usepackage{tabularx}
\usepackage{multirow}
\usepackage[table]{xcolor}
\usepackage{makecell}

\title{SeDeM: Selective Decompression of Hidden-State Memories for Long-Context Question Answering}

\author{
  Maryam Haghifam \and
  Jason Cong \and
  Yizhou Sun \\
  University of California, Los Angeles, USA \\
  \texttt{\{maryamhgf, cong, yzsun\}@cs.ucla.edu}
}

\begin{document}
\maketitle

\begin{figure*}[t]
    \centering
    \includegraphics[width=\textwidth]{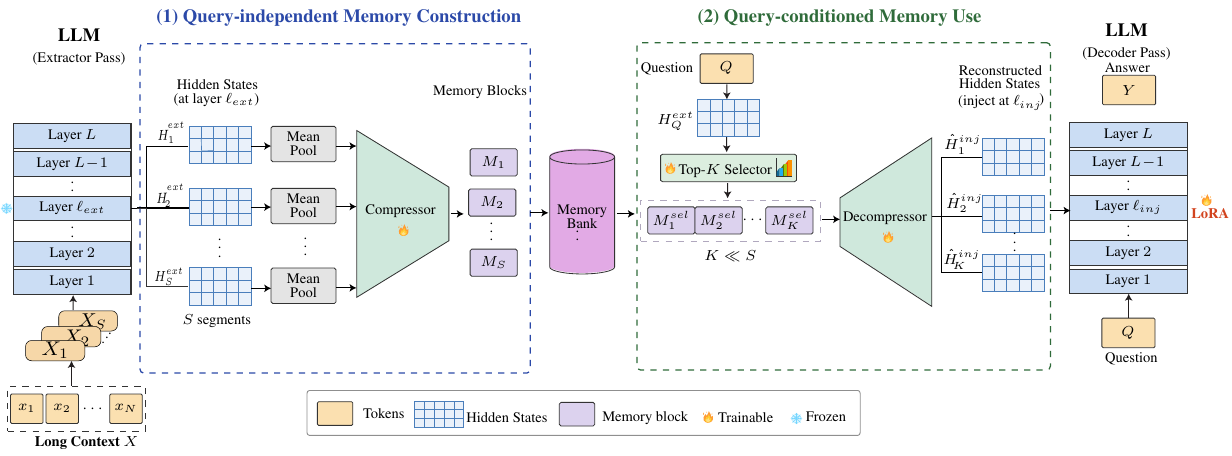}
    \caption{
    Overview of SeDeM. The model stores long-context information in a compact memory bank, selects query-relevant blocks, selectively decompresses them into decoder-compatible hidden states, and injects the reconstructed states into the decoder for answer generation. The extraction layer $\ell_{ext}$ and the injection layer $\ell_{inj}$ need not be equal.
    }
    \label{fig:main_overview}
\end{figure*}

\begin{abstract}
\input{sections/abstract}
\end{abstract}

\input{sections/introduction}

\input{sections/related_work}

\input{sections/method}

\input{sections/training_objective}

\input{sections/experimental_setup}

\input{sections/results}

\input{sections/analysis_ablation}

\input{sections/conclusion}

\input{sections/limitations}

\section*{Acknowledgments}
We thank AMD for providing GPU computing resources that supported portions of the experimental evaluation. This work was partially supported by NSF 2303037, NSF 2312501, NSF 2531008, SRC JUMP 2.0 Center, \href{https://cdsc.ucla.edu/partners/}{CDSC Industry Partners}, Amazon Research Awards, Snapchat, and Google Gifts.

\bibliography{custom}

\appendix
\input{sections/appendix}

\end{document}

%% file: sections/abstract.tex
Long-context inference with large language models (LLMs) is costly: self-attention during prefill scales quadratically with sequence length, and the key-value (KV) cache grows with the number of processed tokens. Larger context windows also do not ensure reliable evidence use. Context compression reduces this cost, but many soft-compression methods use LLMs as compressors and rely on compact memory tokens both to preserve information and to condition the decoder. We propose SeDeM, a selective decompression framework that decouples compact memory storage from decoder conditioning. SeDeM stores context as compact hidden-state memory blocks, selects query-relevant blocks, and decompresses only the selected blocks for decoder conditioning. Thus, the decoder avoids both full-context processing and direct generation from highly compressed memory slots. On four long-context QA benchmarks, SeDeM achieves higher QA scores than the compression baselines in our main comparison in both 1B and 3B same-backbone settings, and with the 3B backbone exceeds full-context fine-tuning on three datasets. %The learned selector uses block-level evidence supervision during training. 
SeDeM also provides favorable quality--efficiency trade-offs, achieving 1.74--2.46$\times$ lower online time-to-first-token and 1.08--1.10$\times$ higher autoregressive decoding throughput relative to ICAE while maintaining strong answer quality.

%% file: sections/introduction.tex
\section{Introduction}
\label{sec:introduction}

Long-context inference with LLMs is costly: self-attention during prefill scales quadratically with sequence length, and the KV cache grows with the number of tokens. Enlarging the context window does not guarantee reliable answer quality in question-answering (QA) tasks: LLMs can fail to use relevant information in long inputs, especially when the evidence is not near the beginning or end of the context~\citep{liu2024lost}. Long-context QA therefore needs methods that are both efficient and selective in how they expose evidence to the decoder.

Context compression addresses the efficiency problem by replacing the full input with a shorter representation before generation. Hard prompt-compression methods prune or select discrete tokens~\citep{li2023selectivecontext,jiang2023llmlingua,jiang2024longllmlingua,pan2024llmlingua2}, while soft compression methods represent the context with continuous memory tokens~\citep{chevalier2023adapting,ge2024icae,li2025generalized,zhang2024activationbeacon}. Many soft-compression methods use the LLM itself as the compressor: learned memory or compression tokens are inserted into the input and updated through self-attention across Transformer layers. Prior work has argued that this design can overwrite information aggregated in earlier layers, causing compact memories to drift away from localized source content~\citep{ye2026comprexit}. Moreover, the same compact slots are often consumed directly by the decoder as a soft prefix, so they must both store source information and serve as decoder-conditioning states. At high compression ratios, compressed memory slots must both preserve source information and provide effective decoder conditioning, while the memories themselves must still be processed through the decoder layers. In our controlled ablations, direct memory conditioning yields lower QA performance than decompressing memories into decoder-compatible hidden states.

We propose SeDeM (\emph{Se}lective \emph{De}compression of Hidden-State \emph{M}emories) as a selective decompression framework for long-context question answering. The framework has two stages: query-independent memory construction and query-conditioned memory use. In the first stage, hidden states from an intermediate encoder layer are stored as compact segment-level memory blocks that together form a memory bank for the context. In the second stage, the query determines which memory blocks are selected, and only those blocks are decompressed into hidden states compatible with an intermediate decoder layer for answer generation.

Three benefits follow from this design. First, the constructed memory bank can be reused across multiple queries over the same context. Second, LLM computation scales with the selection budget rather than the full context length, which can better control the quality-speed trade-off. Third, the design supports cross-model use, where a smaller encoder constructs memories that are decompressed for use by a larger decoder, which separates the cost of context processing from the capacity used for answer generation and adds another level of flexibility.

We evaluate SeDeM on 2WikiMultiHopQA~\citep{ho-etal-2020-constructing},
MuSiQue~\citep{trivedi-etal-2022-musique},
QASPER~\citep{dasigi-etal-2021-dataset}, and
HotpotQA-Distractor~\citep{yang-etal-2018-hotpotqa}. In both Llama-3.2-1B and Llama-3.2-3B same-backbone settings, SeDeM achieves higher scores than the compression baselines in our main comparison across the four datasets. The learned selector is trained with block-level evidence supervision, which several of these baselines do not use; controlled experiments that separate selection from compression are reported in Section~\ref{sec:controls_summary}. With the 3B backbone, SeDeM also surpasses the full-context fine-tuned reference on 2WikiMHQA, QASPER, and HotpotQA-Distractor, while remaining below it on MuSiQue. In a matched comparison with the recent ComprExIT method, SeDeM exposes distinct quality--efficiency operating points rather than uniform superiority (Section~\ref{sec:comprexit_comparison}). Compared with ICAE~\citep{ge2024icae}, SeDeM achieves $1.74\times$--$2.46\times$ lower online TTFT and $1.08\times$--$1.10\times$ higher decoding throughput across the 1B and 3B backbones. Ablations show that direct memory conditioning is substantially weaker than selective decompression, suggesting that the gains come from expanding selected memories into decoder-compatible hidden states rather than from compression alone.

%% file: sections/related_work.tex
\section{Related Work}

\paragraph{Hard prompt compression.}
Hard prompt-compression methods reduce long-context cost by selecting or
pruning discrete input tokens before generation. SelectiveContext~\citep{li2023selectivecontext}
uses language-model surprisal to remove less informative lexical units, while
LLMLingua~\citep{jiang2023llmlingua}, LongLLMLingua~\citep{jiang2024longllmlingua},
and LLMLingua-2~\citep{pan2024llmlingua2} learn or control token-level
compression under a target budget. These methods preserve the standard text interface, but their compressed representation remains a sequence of discrete input tokens. Consequently, reducing the number of positions processed by the LLM requires removing tokens, rather than encoding information from a longer span into fewer learned representations.

\paragraph{Soft and memory-based context compression.}
A second line of work replaces token pruning with continuous memory
representations. Gist tokens~\citep{mu2023gist} and
AutoCompressors~\citep{chevalier2023adapting} train language models to
summarize prompts or segments into soft tokens. ICAE~\citep{ge2024icae}
treats context compression as in-context autoencoding and lets the LLM consume
memory slots directly, while 500xCompressor~\citep{li2025generalized} maps text
into a small number of special memory tokens. CompLLM~\citep{berton2025compllm}
compresses segments independently so that compressed representations can be
reused across queries, and HMT~\citep{he2025hmt} retrieves structured memory
components during long-context processing. Several of these methods follow an
LLM-as-compressor paradigm, where self-attention repeatedly updates compression
tokens. ComprExIT~\citep{ye2026comprexit} argues that compression-token methods can suffer from weak coordination among memory tokens and progressive information loss across Transformer layers. To address this, it directly extracts and transmits information from frozen intermediate hidden states rather than relying solely on learned compression tokens. In
contrast, SeDeM treats memories as compact storage rather than as the
final decoder input: selected memory blocks are decompressed into
decoder-compatible hidden states before generation.

\paragraph{KV-cache and activation compression.}
Another line of work reduces inference cost by modifying the Transformer states
stored or reused during decoding. StreamingLLM~\citep{xiao2024streamingllm}
preserves attention sinks and a recent window, while H2O~\citep{zhang2023h2o},
SnapKV~\citep{li2024snapkv}, PyramidKV~\citep{cai2024pyramidkv},
KIVI~\citep{liu2024kivi}, CacheGen~\citep{liu2024cachegen}, and Activation
Beacon~\citep{zhang2024activationbeacon} reduce cache size, memory bandwidth,
or attention cost over long histories. More recently, ClusterAttn~\citep{zhang2025clusterattn} and RefreshKV~\citep{xu2025refreshkv} compress or update KV states associated with the current inference context. Other methods use semantic units as the basis of compression: ChunkKV~\citep{liu2025chunkkv} selects informative semantic chunks as the basic compression unit, while SemantiCache~\citep{wu2026semanticache} clusters and merges semantically related KV states into compact representations. These approaches are related but potentially complementary to reusable query-independent memory construction. \citet{eyuboglu2026cartridges} take a different route and train a compact, corpus-specific KV cache, termed a Cartridge, offline through self-study and context distillation, amortizing this optimization across subsequent queries to the same corpus. SeDeM instead learns reusable segment-level hidden-state memories, selects query-relevant blocks, and selectively decompresses them before decoder-layer conditioning.

%% file: sections/method.tex
\section{Methodology}
\label{sec:methodology}

Let \(X=(x_1,\ldots,x_N)\) denote a long context with $N$ tokens, \(Q\) a query, and
\(Y\) the target answer. The question answering (QA) problem is then to model \(p(Y \mid X,Q)\). %Standard full-context inference models \(p(Y|X,Q)\) require the decoder to process the full context. 
%Instead, SeDeM constructs query-independent compressed memories from context segments, selects only those relevant ones to \(Q\), and decompresses the selected blocks into decoder-compatible hidden states. These reconstructed states are injected at decoder layer \(\ell_{\mathrm{inject}}\), so answer generation conditions on selected hidden-state reconstructions rather than on the full token sequence or on memory tokens alone.

SeDeM consists of three components: a compressor \(C_{\theta}\), a top-\(K\) selector \(R_{\phi}\), and a decompressor \(D_{\psi}\).  During the compression stage, the long context $X$ is first divided into segments. For each segment, a frozen query-independent encoder extracts hidden states from a chosen Transformer layer \(\ell_{\mathrm{extract}}\), and the compressor maps these states into a compact memory block. During the decoding stage, given a query $Q$, the selector chooses relevant memories, and the decompressor expands only the selected blocks into hidden states compatible with injection at decoder layer \(\ell_{\mathrm{inject}}\), to generate the answer $Y$. The extraction and injection layers are independent empirical hyperparameters rather than being constrained by a fixed relation. Their choice reflects a quality--efficiency trade-off: earlier extraction and later injection reduce computation, while the layer pair also affects reconstruction and downstream QA quality (Appendix~\ref{app:layer_depth_ablation}).

\subsection{The Compressor}
\label{sec:compressor}

For each context segment \(X_s\), we use an LLM encoder as a feature extractor and take hidden states from a chosen extraction layer \(\ell_{\mathrm{extract}}\):
\begin{equation}
    H_s^{(\ell_{\mathrm{extract}})}
    =
    \mathrm{LLM_{Enc}}^{(\ell_{\mathrm{extract}})}(X_s)
    \in
    \mathbb{R}^{T \times d_{\mathrm{enc}}}.
    \label{eq:segment_hidden_states_method}
\end{equation}
where $T$ denotes the number of tokens in a segment and $d_{\mathrm{enc}}$ denotes the encoder hidden-state dimension.
This encoder provides contextualized representations for each token in a segment; compression is then performed by local pooling and projection.
%the actual compression is performed by local pooling and projection, not by inserting learned compression tokens into the LLM self-attention layers.

%For clarity, we describe the compressor 
For a segment of length \(T=N_M C\), where \(N_M\) is the number of memory slots per segment and \(C\) is the per-slot compression factor, the compressor partitions the hidden states into \(N_M\) contiguous chunks of size \(C\). Thus, each memory slot summarizes one local chunk. For each memory slot \(j \in \{1,\ldots,N_M\}\), we compute a local mean-pooled representation:
\begin{equation}
    \bar{h}_{s,j}^{(\ell_{\mathrm{extract}})}
    =
    \frac{1}{C}
    \sum_{k=1}^{C}
    H_s^{(\ell_{\mathrm{extract}})}[(j-1)C+k],
    \label{eq:local_mean_pooling}
\end{equation}
where
\(\bar{h}_{s,j}^{(\ell_{\mathrm{extract}})}
\in \mathbb{R}^{d_{\mathrm{enc}}}\).

Mean pooling gives a parameter-free local compression operator over adjacent
contextualized hidden states. The intuition is that neighboring contextual
states within a short window carry overlapping local information, so their
mean can serve as a compact local summary. This keeps the input
projection independent of the compression factor \(C\). %: the compressor uses \(W_{\mathrm{in}}\in\mathbb{R}^{d_{\mathrm{enc}}\times d_{\mathrm{dec}}}\) rather than a projection from \(C d_{\mathrm{enc}}\) to \(d_{\mathrm{dec}}\).

A shared projection maps each pooled vector into the LLM decoder hidden dimension:
\begin{equation}
    m_{s,j}
    =
    \bar{h}_{s,j}^{(\ell_{\mathrm{extract}})}
    W_{\mathrm{in}} .
    \label{eq:input_projection}
\end{equation}
where \(W_{\mathrm{in}}
\in \mathbb{R}^{d_{\mathrm{enc}} \times d_{\mathrm{dec}}}\) and \(m_{s,j} \in \mathbb{R}^{d_{\mathrm{dec}}}\).
%The combine design  uses \(W_{\mathrm{in}}\in\mathbb{R}^{d_{\mathrm{enc}}\times d_{\mathrm{dec}}}\) rather than a projection from \(C d_{\mathrm{enc}}\) to \(d_{\mathrm{dec}}\).

The resulting memory block for segment \(s\) is then a concatenation of $N_M$ local chunks:
\begin{equation}
    M_s
    =
    \left[
    m_{s,1};
    \ldots;
    m_{s,N_M}
    \right]
    \in
    \mathbb{R}^{N_M \times d_{\mathrm{dec}}}.
    \label{eq:segment_memory_method}
\end{equation}

The compressor therefore reduces each segment from \(T\) states to \(N_M\) memory vectors and aligns them with the decoder hidden dimension using a shared linear projection. It has no token-to-token attention: each memory vector is obtained by local pooling followed by the shared projection \(W_{\mathrm{in}}\). Furthermore, unlike compression-token approaches, the compressor does not add extra tokens to the LLM self-attention layers.

\subsection{The Top-K Selector over Memory Blocks}
\label{sec:retrieval}
Our compression is query-independent; the query only affects scoring and selection. The segment memories form a global memory bank \(M_{\mathrm{bank}}=\mathrm{Concat}_{s=1}^{S}M_s\in\mathbb{R}^{(S N_M)\times d_{\mathrm{dec}}}\). We partition this bank into \(N_{\mathrm{blk}}\) contiguous memory blocks of size \(N_B\), where \(N_{\mathrm{blk}}=SN_M/N_B\), and define \(B_n=M_{\mathrm{bank}}[(n-1)N_B+1:nN_B]\). Each block satisfies \(B_n\in\mathbb{R}^{N_B\times d_{\mathrm{dec}}}\). In our main setting, \(N_B=N_M\), so each memory block corresponds to one context segment.

We use the term selection rather than text retrieval because the model selects
latent memory blocks, not raw text. Unlike retrieval-augmented generation
(RAG)~\citep{lewis2020retrieval}, it does not append retrieved passages to the prompt; selected memories
are later decompressed into decoder-layer hidden states, which can significantly reduce the inference time compared to retrieving and processing raw context.

To score memory blocks, we encode the query with the same encoder up to
\(\ell_{\mathrm{extract}}\), obtaining
\(H_Q^{(\ell_{\mathrm{extract}})}
=\mathrm{LLM}_{\mathrm{Enc}}^{(\ell_{\mathrm{extract}})}(Q)
\in \mathbb{R}^{N_q \times d_{\mathrm{enc}}}\).
Using the same extraction depth keeps query features aligned with the encoder
states used to construct the memory bank. Learned projections then map query
states and memory slots into a shared scoring space.

The selector module uses \(R\) learned scoring heads. For head \(r\), query token \(n_q\), memory block \(B_n\), and memory slot \(n_b\), we compute
\begin{equation}
\begin{aligned}
q_{r,n_q} &= \mathrm{LN}\!\left(H_Q^{(\ell_{\mathrm{extract}})}[n_q]\right)W_Q^{(r)},\\
k_{r,n,n_b} &= \mathrm{LN}\!\left(B_n[n_b]\right)W_K^{(r)},\\
\alpha_{r,n_q,n,n_b} &= \cos(q_{r,n_q},k_{r,n,n_b}).
\end{aligned}
\label{eq:retrieval_similarity_method}
\end{equation} where $\mathrm{LN}(\cdot)$ denotes layer normalization.

Following ColBERT-style late interaction~\citep{khattab2020colbert}, for each
query token, MaxSim retains the strongest matching memory slot within block
\(B_n\):
\begin{equation}
\alpha^{\max}_{r,n_q,n}
=
\max_{n_b\in\{1,\ldots,N_B\}}
\alpha_{r,n_q,n,n_b}.
\label{eq:maxsim}
\end{equation}

The score \(s_n\) for block \(B_n\) is computed by summing MaxSim similarities over query tokens and averaging over scoring heads:
\begin{equation}
    s_n
    =
    \frac{1}{R}
    \sum_{r=1}^{R}
    \sum_{n_q=1}^{N_q}
    \alpha^{\max}_{r,n_q,n}.
    \label{eq:block_score_method}
\end{equation}

The model ranks all memory blocks by their scores and selects the \(K\) highest-scoring blocks. We denote the selected block indices by
\(\mathcal{I}(Q)\), with \(|\mathcal{I}(Q)|=K\).

This late-interaction score leverages fine-grained token-to-slot matching before block-level aggregation. Different query tokens can therefore match different memory slots within the same block, rather than relying on a single pooled query or block representation.

\subsection{The Decompressor}
\label{sec:decompressor}
The decompressor is a shared two-layer MLP applied independently to each memory vector in the selected blocks. In our main setting, where \(N_B=N_M\), each selected block corresponds to one context segment. Given a memory vector \(m_{s,j}\) from a selected segment \(s\), with \(1 \leq j \leq N_M\), the decompressor computes
\begin{equation}
\begin{aligned}
    a_{s,j}
    &=
    \mathrm{GELU}
    \left(
        \mathrm{LN}(m_{s,j})W_1^{\mathrm{dec}}
        +
        b_1^{\mathrm{dec}}
    \right),\\
    z_{s,j}
    &=
    a_{s,j}W_2^{\mathrm{dec}} + b_2^{\mathrm{dec}},
\end{aligned}
\label{eq:decompressor_mlp}
\end{equation}
where \(z_{s,j}\in\mathbb{R}^{C d_{\mathrm{dec}}}\), corresponding to \(C\) decoder hidden states of dimension \(d_{\mathrm{dec}}\).

We reshape \(z_{s,j}\) into \(C\) decoder-space hidden states
\(\widehat{H}_{s,j}\in\mathbb{R}^{C\times d_{\mathrm{dec}}}\). Concatenating
over \(j=1,\ldots,N_M\) gives a segment-level reconstruction
\(\widehat{H}_s\in\mathbb{R}^{T\times d_{\mathrm{dec}}}\), and concatenating
over the segments corresponding to the selected blocks \(\mathcal{I}(Q)\) gives
\(\widehat{H}_{\mathcal{I}(Q)}\in\mathbb{R}^{KT\times d_{\mathrm{dec}}}\).

For answer generation, the LLM decoder first processes the query-side prefix up to
injection layer \(\ell_{\mathrm{inject}}\). At this layer, we prepend
\(\widehat{H}_{\mathcal{I}(Q)}\) to the query hidden states. The decoder then
continues from layer \(\ell_{\mathrm{inject}}+1\) to \(L\) over the augmented
sequence and generates the answer autoregressively.

Above the injection layer, the decoder processes approximately the query-side
prefix together with \(KT\) reconstructed positions, rather than the full
\(N\)-token context.

The decompressor is trained to produce hidden states compatible with the decoder
layer at which they are consumed, rather than to reconstruct text. This keeps
the remaining decoder layers closer to the activation regime they would see
when processing the selected context directly.

%sparse attention

%% file: sections/training_objective.tex
\section{Training Objective}
\label{sec:training_objective}
Training proceeds in two stages. Stage~1 trains the compressor and decompressor using context-window next-token reconstruction, hidden-state reconstruction, and decoder distillation from a raw-context teacher path. Stage~2 trains the full query-conditioned pipeline, including selection, selective decompression, hidden-state injection, and answer generation. Additional optimization details are provided in Appendix~\ref{app:training_configuration}. In the same-backbone setting, the LLM uses frozen pretrained weights, while Stage~2 trains lightweight LoRA adapters on the shared backbone for task adaptation.

\paragraph{Stage 1: Reconstruction pretraining.}
Only the compressor and decompressor are trained in this stage; the decoder LLM is used to define reconstruction and distillation losses, while its pretrained weights remain frozen. Let
\(\widetilde{H}_s =
H_s^{(\ell_{\mathrm{extract}})} W_{\mathrm{in}}
\in \mathbb{R}^{T \times d_{\mathrm{dec}}}\)
denote the encoder hidden states projected into the decoder hidden dimension.

This projected representation serves as an alignment target for hidden-state
reconstruction. The context-window next-token loss, distillation loss, and
Stage~2 QA loss further adapt the reconstructed states to be useful for decoder
conditioning.

Using the same compression factor \(C\) as in the compressor, let
\(\operatorname{Pool}_C(\cdot)\) denote non-overlapping mean pooling over
windows of size \(C\). We define the pooled reconstruction and target as
\(\bar{\widehat{H}}_s=\operatorname{Pool}_C(\widehat{H}_s)\) and
\(\bar{\widetilde{H}}_s=\operatorname{Pool}_C(\widetilde{H}_s)\).

The reconstruction objective combines token-level and pooled directional
alignment:
\begin{equation}
    \mathcal{L}_{\mathrm{dir}}(s)
    =
    1 -
    \frac{1}{T}
    \sum_{t=1}^{T}
    \cos\!\left(
        \widehat{H}_{s,t},
        \widetilde{H}_{s,t}
    \right).
    \label{eq:loss_dir}
\end{equation}
\begin{equation}
    \mathcal{L}_{\mathrm{pool}}(s)
    =
    1 -
    \frac{1}{N_M}
    \sum_{j=1}^{N_M}
    \cos\!\left(
        \bar{\widehat{H}}_{s,j},
        \bar{\widetilde{H}}_{s,j}
    \right).
    \label{eq:loss_pool}
\end{equation}
and
\begin{equation}
    \mathcal{L}_{\mathrm{rec}}
    =
    \mathbb{E}_{s}
    \left[
        \mathcal{L}_{\mathrm{dir}}(s)
        +
        \gamma \mathcal{L}_{\mathrm{pool}}(s)
    \right].
    \label{eq:loss_rec}
\end{equation}

The directional term aligns reconstructed hidden states with projected encoder states, while the pooled term preserves local summaries at the memory-slot scale.

Stage~1 also applies a context-window next-token loss \(\mathcal{L}_{\mathrm{ctx}}\) by passing the reconstructed states through the frozen decoder. We further use a distillation loss \(\mathcal{L}_{\mathrm{distill}}\). The frozen decoder on the raw context provides the teacher distribution, while the same decoder conditioned on reconstructed states provides the student distribution. Over the continuation positions, we minimize the token-averaged KL divergence from the teacher distribution to the student distribution, with both distributions computed at temperature \(\tau\).
The Stage~1 objective is
\begin{equation}
    \mathcal{L}^{(1)}
    =
    \mathcal{L}_{\mathrm{ctx}}
    +
    \lambda_{\mathrm{distill}}
    \mathcal{L}_{\mathrm{distill}}
    +
    \lambda_{\mathrm{rec}}^{(1)}
    \mathcal{L}_{\mathrm{rec}} .
    \label{eq:loss_stage1}
\end{equation}
%We use \(\lambda_{\mathrm{distill}}=0.5\) and \(\lambda_{\mathrm{rec}}^{(1)}=1.0\).

\paragraph{Stage 2: Selection-supervised task training.}
In Stage~2, we train the selection module, continue updating the compressor and decompressor, and activate the decoder LLM LoRA adapters. For each training example \((X,Q,Y)\), the selector selects \(\mathcal{I}(Q)\). The selected memories are decompressed into
\(\widehat{H}_{\mathcal{I}(Q)}\) and injected into the decoder at layer \(\ell_{\mathrm{inject}}\). The QA generation
loss is
\begin{equation}
    \mathcal{L}_{\mathrm{LM}}
    =
    -
    \frac{1}{N_{\mathrm{ans}}}
    \sum_{t=1}^{N_{\mathrm{ans}}}
    \log
    p\!\left(
        y_t
        \mid
        y_{<t},
        Q,
        \widehat{H}_{\mathcal{I}(Q)}
    \right).
    \label{eq:loss_lm}
\end{equation}

Because Top-\(K\) selection is discrete, \(\mathcal{L}_{\mathrm{LM}}\) does not directly optimize the ranking scores for unselected blocks. We therefore add a selection loss \(\mathcal{L}_{\mathrm{ret}}\), defined in Appendix~\ref{app:retrieval_supervision}, using block-level evidence labels.

The Stage~2 objective is
\begin{equation}
    \mathcal{L}^{(2)}
    =
    \mathcal{L}_{\mathrm{LM}}
    +
    \lambda_{\mathrm{ret}}
    \mathcal{L}_{\mathrm{ret}}
    +
    \lambda_{\mathrm{rec}}^{(2)}
    \mathcal{L}_{\mathrm{rec}}.
    \label{eq:loss_total}
\end{equation}
We retain \(\mathcal{L}_{\mathrm{rec}}\) during Stage~2 as a regularizer over all encoded memory blocks. This keeps the reconstructed states close to the projected encoder states while \(\mathcal{L}_{\mathrm{LM}}\) adapts them for answer generation.

%% file: sections/experimental_setup.tex
\section{Experimental Setup}
\label{sec:experiments}

\paragraph{Datasets.}
We evaluate on four long-context QA benchmarks with different context structures and
hop depths: HotpotQA-Distractor, 2WikiMultiHopQA,
MuSiQue, and QASPER (Table~\ref{tab:datasets}, Appendix~\ref{app:dataset_processing}). We segment each context before
compression and map available evidence annotations to segment-level labels,
which are used only to supervise the selection loss during training. Dataset sources, splits, and preprocessing details are provided in Appendix~\ref{app:dataset_processing}.

\paragraph{Models and compression configuration.}
We use Llama-3.2-Base models~\citep{meta2024llama32,grattafiori2024llama3} as backbones to isolate
the effect of the compression mechanism from instruction-tuning behavior. We
evaluate same-model settings with Llama-3.2-1B-Base and Llama-3.2-3B-Base, and a
cross-model setting with a 1B LLM encoder and 3B LLM decoder. We use segment length $T=128$,
compression factor $C=4$. This corresponds to a \(4\times\) storage compression ratio: each
128-token segment is stored as 32 memory slots. The selection budget
\(K\) controls the number of reconstructed decoder-conditioning
positions, \(128K\), and is not included in the storage compression
ratio. The selector uses $R=4$ late-interaction heads. We set $K=2$ for HotpotQA-Distractor and 2WikiMHQA, $K=4$ for MuSiQue, and $K=8$ for QASPER.

\paragraph{Baselines.}
We compare against full-context frozen and fine-tuned with LoRA references and
representative compression baselines on the same downstream train/validation
splits and Llama-3.2 backbones. The compression baselines
include LongLLMLingua at $4\times$ compression, ICAE, HMT, Activation Beacon, and 500xCompressor. Reproduction details and deviations from the original
baseline settings are reported in Appendix~\ref{app:baseline_reproduction}.

\paragraph{Metrics.}
We report token-level F1 with SQuAD-style normalization and ROUGE-L F1 on a
0--100 scale. Exact match (EM) is reported for SeDeM in Table~\ref{tab:sedem_em}, the operating-point comparison
in Table~\ref{tab:comprexit_operating_points}, and in
Appendix~\ref{app:k_sweep}. For efficiency, we report online TTFT and
autoregressive decoding throughput, with no document-side caching or
amortization. Storage compression, decoder-conditioning length, FLOPs, and
measured TTFT are distinct quantities and are reported separately.

%% file: sections/results.tex
\input{sections/main_table}

\section{Results}
\label{sec:results}
We first report the main QA results and the matched quality--efficiency
comparison, then evaluate inference efficiency and long-context length
behavior, followed by ablations and representation analyses.

\subsection{Main Long-Context QA Results}
\label{sec:main_results}
Table~\ref{tab:main_qa_results} reports the main long-context QA results.
Exact-match (EM) results for SeDeM are reported in Table~\ref{tab:sedem_em}. In the Llama-3.2-1B setting, SeDeM achieves higher scores than all compression baselines in Table~\ref{tab:main_qa_results} across the four datasets and also improves over the full-context fine-tuned reference. The gains are especially large on 2WikiMHQA and HotpotQA-Distractor, where selective decompression provides substantially more useful context than the compressed-memory baselines. A matched comparison with the recent ComprExIT method, which is not included in Table~\ref{tab:main_qa_results}, is reported separately in Section~\ref{sec:comprexit_comparison} and shows a quality--efficiency trade-off rather than uniform superiority.

In the Llama-3.2-3B setting, SeDeM again achieves higher scores than all compression baselines in Table~\ref{tab:main_qa_results} across the four datasets. Compared with the full-context fine-tuned reference, it improves on 2WikiMHQA, QASPER, and HotpotQA-Distractor, but remains lower on MuSiQue. This suggests that selective decompression is especially effective when the selected compressed evidence is sufficient for answer generation, while full-context fine-tuning can remain stronger on some multi-hop settings that benefit from broader context access.

The cross-model setting uses a Llama-3.2-1B encoder with a Llama-3.2-3B decoder. This setting tests whether memories constructed by a smaller encoder can be consumed by a larger decoder after projection and decompression. The cross-model results remain strong on 2WikiMHQA and HotpotQA-Distractor, showing that the compressed hidden-state representation is not restricted to identical encoder--decoder backbones. On 2WikiMHQA, the cross-model result of \(63.44\) exceeds the 3B full-context fine-tuned reference of \(62.50\), but remains below the 3B same-model SeDeM result of \(67.25\). Section~\ref{sec:ruler_qa2} evaluates length behavior on RULER \texttt{qa\_2} at 4K, 8K, and 16K contexts.

\subsection{Matched Comparison with ComprExIT: Quality--Efficiency Operating Points}
\label{sec:comprexit_comparison}

ComprExIT~\citep{ye2026comprexit} is the most closely related recent hidden-state compression method, so we report a matched comparison using its official implementation, its released NTP-pretrained checkpoint, our dataset-specific SFT data, complete validation sets, Llama-3.2-1B, and a \(4\times\) compression ratio. Because the better ComprExIT configuration is dataset-dependent, we report both a reference 512-token configuration, following ComprExIT's published context length, and a 128-token control matched to SeDeM's segment granularity, and compare quality against the per-dataset best configuration. Table~\ref{tab:comprexit_operating_points} reports F1 and exact match (EM) for the two SeDeM operating points and both ComprExIT configurations.

\begin{table}[t]
\centering
\small
\begin{tabular}{@{}lrrr@{}}
\toprule
Configuration & F1 & EM & TTFT (ms) \\
\midrule

\multicolumn{4}{c}{HotpotQA-Distractor} \\
\cmidrule(lr){1-4}
ComprExIT (reference)   & 55.50 & 40.74 & 111.7 \\
ComprExIT (matched)     & 54.27 & 39.39 & 373.8 \\
SeDeM learned-\(K{=}2\) & 50.93 & 38.42 & 47.43 \\
SeDeM full-bank         & 59.17 & 45.05 & 50.01 \\

\midrule
\multicolumn{4}{c}{2WikiMHQA} \\
\cmidrule(lr){1-4}
ComprExIT (reference)   & 58.76 & 51.76 & 110.7 \\
ComprExIT (matched)     & 61.74 & 54.42 & 377.9 \\
SeDeM learned-\(K{=}2\) & 55.69 & 49.40 & 33.49 \\
SeDeM full-bank         & 67.28 & 61.36 & 49.81 \\

\bottomrule
\end{tabular}
\caption{
Matched ComprExIT comparison with Llama-3.2-1B at \(4\times\) compression. The reference configuration follows ComprExIT's published 512-token context length; the matched control uses SeDeM's 128-token segment granularity. Full-bank SeDeM disables selection and decompresses all memory blocks. The HotpotQA-Distractor comparison is supervision-matched, whereas 2WikiMHQA uses additional evidence-related Stage-2 supervision unavailable to ComprExIT.
}
\label{tab:comprexit_operating_points}
\end{table}

The selective learned-\(K{=}2\) configuration scores \(4.57\) F1 below the
best ComprExIT configuration on HotpotQA-Distractor and \(6.05\) F1 below
it on 2WikiMHQA, while full-bank SeDeM scores above ComprExIT on both
datasets. Under the matched A100 protocol, learned-\(K{=}2\) and
full-bank SeDeM obtain TTFTs of \(47.43/50.01\) ms on
HotpotQA-Distractor and \(33.49/49.81\) ms on 2WikiMHQA, respectively,
compared with \(111.7\)--\(373.8\) ms and \(110.7\)--\(377.9\) ms for
the two ComprExIT configurations.

The relatively small first-token latency increase from full-bank
decompression arises because context compression and query-side processing
account for a substantial portion of SeDeM's online TTFT and do not scale
with the number of retained blocks. Increasing the selection budget
primarily affects decompression and decoder-side computation. The
2WikiMHQA full-bank model used evidence-related Stage-2 supervision
unavailable to ComprExIT, so its quality comparison is not
supervision-matched.

\subsection{Inference Efficiency}
\label{sec:inference_efficiency}

Table~\ref{tab:efficiency_varb_icae} in Appendix~\ref{app:efficiency_overhead} compares inference efficiency against ICAE using TTFT and decoding throughput. On average across the four datasets, SeDeM reduces online TTFT by \(1.74\times\) with the 1B backbone and \(2.46\times\) with the 3B backbone, and improves decoding throughput by \(1.08\times\) and \(1.10\times\), respectively. For our method, TTFT includes online compression, selection, and decoder prefill before the first generated answer token.

Two implementation properties reduce TTFT: segment-level compression runs in parallel across segments and both compression and query encoding terminate at \(\ell_{\mathrm{extract}}\). For example, in the 3B configuration, early-exit query encoding reduces selection latency from approximately \(26\) ms to \(13\) ms. Implementation details are deferred to Appendix~\ref{app:implementation_details}. Our added inference-time modules correspond to \(3.82\%\) and \(3.49\%\) of the frozen 1B and 3B backbones, respectively; detailed parameter and FLOP information is provided in Appendix~\ref{app:inference_overhead_app}.

\paragraph{Effective-rank analysis.}
Effective-rank analysis in Appendix~\ref{app:erank_definition} shows higher
normalized effective rank for SeDeM than for ICAE on all four datasets
(average \(78.7\%\) vs.\ \(52.3\%\)), suggesting less memory-slot collapse.

\subsection{Length Behavior on RULER \texttt{qa\_2}}
\label{sec:ruler_qa2}

We evaluate SeDeM on RULER's \texttt{qa\_2} task~\citep{hsieh2024ruler}, a
multi-document multi-hop QA task derived from HotpotQA, at 4K, 8K, and 16K
context lengths, using the official \texttt{string\_match\_part} metric (500
examples per length, greedy decoding, \texttt{max\_new\_tokens}=128). All
methods use the Llama-3.2-3B-Base backbone, and SeDeM keeps its
main-experiment configuration without task-specific tuning.

\begin{table}[t]
\centering
\small
\setlength{\tabcolsep}{5pt}
\renewcommand{\arraystretch}{0.95}
\begin{tabular}{@{}lccc@{}}
\toprule
Method & 4K & 8K & 16K \\
\midrule
Full-context reference & 49.4 & 48.8 & 49.4 \\
SeDeM                  & 31.6 & 27.2 & 22.6 \\
ICAE                   & 11.2 & 13.0 & \phantom{0}7.6 \\
\bottomrule
\end{tabular}
\caption{RULER \texttt{qa\_2} results (\texttt{string\_match\_part}, 500
examples per length, Llama-3.2-3B-Base). SeDeM remains stronger than ICAE
through 16K, but its score declines as context length increases.}
\label{tab:ruler_qa2}
\end{table}

Table~\ref{tab:ruler_qa2} shows that SeDeM retains approximately \(64\%\),
\(56\%\), and \(46\%\) of the full-context score at 4K, 8K, and 16K,
respectively, compared with \(23\%\), \(27\%\), and \(15\%\) for ICAE. SeDeM
therefore remains stronger than ICAE through 16K, but its performance decreases
as context length increases, so we do not claim length invariance; evaluation
beyond 16K remains future work.

%% file: sections/main_table.tex
\begin{table*}[t]
\centering
\small
\setlength{\tabcolsep}{3pt}
\renewcommand{\arraystretch}{0.95}
\begin{tabular*}{\textwidth}{@{\extracolsep{\fill}}lccccccccc@{}}
\toprule
\multirow{2}{*}{Method}
& \multirow{2}{*}{Category}
& \multicolumn{2}{c}{2Wiki}
& \multicolumn{2}{c}{MuSiQue}
& \multicolumn{2}{c}{QASPER}
& \multicolumn{2}{c}{HotpotQA-Dist.} \\
\cmidrule(lr){3-4}
\cmidrule(lr){5-6}
\cmidrule(lr){7-8}
\cmidrule(lr){9-10}
& & F1 & ROUGE-L & F1 & ROUGE-L & F1 & ROUGE-L & F1 & ROUGE-L \\
\midrule

\multicolumn{10}{c}{\textit{Llama-3.2-1B-Base}} \\
\midrule
Backbone (Frozen)
& Ref.
& 13.39 & 13.35
& 5.77 & 5.68
& 13.09 & 11.61
& 17.54 & 17.43 \\

Backbone (Fine-Tuned)
& Ref.
& 49.36 & 49.28
& 16.96 & 16.67
& 21.42 & 19.58
& 29.16 & 22.92 \\

LongLLMLingua
& Hard
& 22.72 & 22.67
& 12.52 & 12.35
& 17.58 & 15.92
& 28.15 & 28.05 \\

HMT
& Soft
& 27.61 & 27.56
& 4.56 & 4.48
& 13.86 & 13.21
& 27.79 & 27.73 \\

ICAE
& Soft
& 30.87 & 30.81
& 8.55 & 8.45
& 16.88 & 15.88
& 26.79 & 26.71 \\

500xCompressor
& Soft
& 35.83 & 35.79
& 6.70 & 6.64
& 15.86 & 15.05
& 30.99 & 30.81 \\

Activation Beacon
& KV/Act.
& 17.98 & 17.97
& 8.46 & 8.35
& 19.76 & 18.74
& 39.73 & 39.57 \\

\rowcolor{green!12}
SeDeM
& Sel. Soft
& 55.69 & 55.66
& 18.63 & 18.55
& 23.82 & 22.88
& 50.93 & 50.82 \\

\midrule
\multicolumn{10}{c}{\textit{Llama-3.2-3B-Base}} \\
\midrule
Backbone (Frozen)
& Ref.
& 18.30 & 18.26
& 9.50 & 9.38
& 16.73 & 15.17
& 23.06 & 22.95 \\

Backbone (Fine-Tuned)
& Ref.
& 62.50 & 62.49
& 32.54 & 32.29
& 23.44 & 21.50
& 36.73 & 36.56 \\

LongLLMLingua
& Hard
& 27.19 & 27.12
& 19.49 & 19.34
& 23.42 & 21.45
& 39.59 & 39.46 \\

HMT
& Soft
& 34.50 & 34.47
& 6.03 & 5.98
& 15.79 & 15.40
& 31.83 & 31.76 \\

ICAE
& Soft
& 38.77 & 38.70
& 14.00 & 13.88
& 20.21 & 19.09
& 37.99 & 37.82 \\

500xCompressor
& Soft
& 52.14 & 52.09
& 6.98 & 6.92
& 13.73 & 13.06
& 39.15 & 38.96 \\

Activation Beacon
& KV/Act.
& 25.31 & 25.27
& 19.36 & 19.24
& 23.16 & 22.50
& 45.09 & 44.99 \\

\rowcolor{green!12}
SeDeM
& Sel. Soft
& 67.25 & 67.14
& 21.85 & 21.68
& 26.74 & 25.74
& 58.30 & 58.16 \\

\midrule
\multicolumn{10}{c}{\textit{Cross-model: 1B encoder $\rightarrow$ 3B decoder}} \\
\midrule
\rowcolor{green!12}
SeDeM
& Sel. Soft
& 63.44 & 63.31
& 19.24 & 18.03
& 20.23 & 19.48
& 44.39 & 44.21 \\

\bottomrule
\end{tabular*}
\caption{
Main long-context QA results. We report token-level F1 and ROUGE-L; higher is better. Category labels denote full-context references (Ref.), hard prompt compression (Hard), soft memory compression (Soft), selective soft memory compression (Sel. Soft), and KV-cache or activation compression (KV/Act.). The cross-model row uses a 1B encoder and a 3B decoder.
}
\label{tab:main_qa_results}
\end{table*}

\begin{table}[t]
\centering
\small
\begin{tabular}{@{}lrrrr@{}}
\toprule
Backbone & 2Wiki & MuSiQue & QASPER & HotpotQA \\
\midrule
Llama-3.2-1B & 49.40 & 8.56  & 11.11 & 38.42 \\
Llama-3.2-3B & 59.56 & 13.10 & 11.75 & 45.90 \\
\bottomrule
\end{tabular}
\caption{Exact match (EM) for SeDeM in the same-backbone setting.}
\label{tab:sedem_em}
\end{table}

%% file: sections/analysis_ablation.tex
\section{Ablations and Analysis}
\label{sec:ablations}

This section tests the main design choices behind SeDeM: decompression,
block-structured compression, selection budget, decoder adaptation, and
zero-shot transfer. Layer-depth diagnostics, a stability-based view of
intermediate-layer injection, the out-of-distribution transfer test, and the
decoder-adaptation ablation are provided in
Appendices~\ref{app:layer_depth_ablation}, \ref{app:stability_layer_injection},
\ref{app:ood_transfer}, and~\ref{app:lora_ablation}. Unless otherwise stated, selector-disabled ablations disable the selector, omit
\(\mathcal{L}_{\mathrm{ret}}\), and condition the decoder on all segment-level
memory blocks or their reconstructions. This isolates the compression and
decompression pathway from selection quality.

\subsection{Is Decompression Necessary?}
\label{sec:decompressor_ablation}
We evaluate whether the decompression is necessary by replacing selective decompression with direct memory conditioning. In this variant, compressed memory slots are mapped into the decoder embedding space, prepended to the decoder input as soft memory tokens, and processed through the complete decoder; there is no intermediate-layer reconstruction or injection.

Because the decompressor is absent, objectives defined through reconstructed states (hidden-state reconstruction, token-level reconstruction, and distillation) are unavailable. The ablated variants are therefore trained with the next-token loss in Stage~1 and the QA loss in Stage~2, and both stages are retrained from scratch. We evaluate a linear compressor (our default compressor) and a higher-capacity two-layer MLP compressor to test whether additional memory-writer capacity can compensate for removing decompression.

Table~\ref{tab:decompressor_ablation} shows that direct memory conditioning degrades performance. The higher-capacity MLP compressor does not recover the loss, suggesting that the improvement is not simply due to a more expressive memory writer. The results support the role of decompression.

Layer-depth diagnostics in Appendix~\ref{app:layer_depth_ablation} show that decoder injection depth has a large effect in a controlled selector-disabled setting,
consistent with aligning decompressed states to the decoder layer where they are
consumed.

\begin{table}[t]
\centering
\small
\setlength{\tabcolsep}{3pt}
\renewcommand{\arraystretch}{0.95}
\begin{tabular}{@{}lcc@{}}
\toprule
Method & QASPER & HotpotQA-Dist. \\
\midrule
No decomp. (lin.) & 18.06 &  29.69 \\
No decomp. (MLP)  & 18.30 & 30.13 \\
\textbf{SeDeM} & \textbf{26.74} & \textbf{49.39} \\
\bottomrule
\end{tabular}
\caption{
Decompression ablation using Llama-3.2-3B-Base. Scores are token-F1;
ROUGE-L follows the same pattern. The SeDeM reference in this
controlled ablation uses the internal non-final extraction-layer
configuration; Table~\ref{tab:main_qa_results} reports the designated main
configuration.
}
\label{tab:decompressor_ablation}
\end{table}

% \begin{table}[t]
% \centering
% \small
% \setlength{\tabcolsep}{3pt}
% \renewcommand{\arraystretch}{1.05}
% \begin{tabular}{@{}l*{4}{>{\centering\arraybackslash}p{0.80cm}}@{}}
% \toprule
% \multirow{2}{*}{Method}
% & \multicolumn{2}{c}{QASPER}
% & \multicolumn{2}{c}{HotpotQA-Dist.} \\
% \cmidrule(lr){2-3}
% \cmidrule(lr){4-5}
% & F1 & R-L & F1 & R-L \\
% \midrule

% No decomp. (linear)
% & 18.30 & 17.62 & 30.13 & 30.07 \\

% No decomp. (MLP)
% & 18.08 & 17.50 & 29.69 & 29.69 \\

% \textbf{Ours}
% & \textbf{26.74} & \textbf{25.74}
% & \textbf{46.10} & \textbf{45.81} \\

% \bottomrule
% \end{tabular}
% \caption{
% Ablation of the decompression pathway using Llama-3.2-3B-Base on QASPER and HotpotQA-Dist. Directly conditioning the decoder on compressed memory slots degrades performance, even with a higher-capacity MLP compressor.
% }
% \label{tab:decompressor_ablation}
% \end{table}
\subsection{Selection and Memory Structure}
\paragraph{Block-structured memories support selection.}
We also test a learnable-query compressor that attends globally over each
segment instead of assigning memory slots to local contiguous chunks. This more
flexible compressor performs best when all blocks are retained, suggesting that
global slot mixing is less compatible with top-\(K\) selection. Details are
provided in Appendix~\ref{app:ablation_query_compressor}.

\paragraph{Selection budget.}
Table~\ref{tab:k_sweep} in Appendix~\ref{app:k_sweep} summarizes the effect of
the selection budget \(K\). HotpotQA-Distractor is best with a tight budget, while
2WikiMHQA benefits from broader selection before saturating. This shows that \(K\) controls a quality--context trade-off over selected memory blocks, rather than directly corresponding to the number of annotated supporting paragraphs.

\subsection{Disentangling Selection from Compression}
\label{sec:controls_summary}

Because the learned selector uses block-level evidence supervision unavailable
to several compression baselines, Appendix~\ref{app:controls} reports controls
separating selection from compression. Under identical gold-selected input on
HotpotQA-Distractor with the 3B backbone, SeDeM exceeds ICAE by \(4.92\) F1
(\(64.95\) vs.\ \(60.03\)), indicating that the compression--decompression
pathway contributes beyond selection. Matched raw-text RAG is stronger in
answer quality at \(K{=}2\) (\(66.31\) vs.\ \(58.30\) F1), reflecting a
quality--efficiency trade-off rather than superiority over retrieval.
Boundary shifts cause only moderate degradation (\(0.81\)--\(2.14\) F1). Full
supervision and full-bank controls are reported in
Appendix~\ref{app:controls}.

\paragraph{Compression granularity.}
We analyze segment length and compression factor in Appendix~\ref{app:codec_granularity_ablation}. Stronger compression
monotonically reduces quality, while increasing segment length helps up to a
moderate range before saturating.

%what if there is no gold span?

% decribe icae training

%with retreival and this stateless comrpession, we know whcih memory and which part of the text was excatly used!

%% file: sections/conclusion.tex
\section{Conclusion}

We introduced SeDeM, a selective decompression framework for
long-context question answering. SeDeM stores context segments as
compact hidden-state memories, selects query-relevant blocks, and decompresses
only selected blocks into decoder-compatible hidden states. Across four
long-context QA benchmarks, SeDeM achieves higher scores than the
compression baselines in our main comparison in both 1B and 3B same-backbone
settings, while reducing online time-to-first-token and improving decoding
throughput relative to ICAE. A matched comparison with ComprExIT shows that
SeDeM separates compact reusable storage from decoder conditioning and exposes
distinct quality--efficiency operating points: the selective learned-\(K\)
configuration favors latency, while the full-bank configuration favors
quality. Controlled experiments indicate that the compressor--decompressor
pathway contributes beyond supervised selection.
Ablations show that direct memory conditioning is substantially weaker than selective decompression, supporting the separation between compact memory
storage and decoder-layer conditioning.

%% file: sections/limitations.tex
\section*{Limitations}

This work has several limitations. First, our selection loss is trained using
block-level evidence labels, which all four benchmarks provide but several of
the compression baselines in our main comparison do not use. The uncontrolled
main benchmark comparison therefore does not by itself isolate compression
quality from supervised selection; the controlled experiments in
Section~\ref{sec:controls_summary} and Appendix~\ref{app:controls} are
provided to separate retrieval from compression and decompression, and this
supervision difference should be kept in mind when interpreting the main
tables. When such labels are unavailable, SeDeM can still be operated in the
selector-disabled mode used in our ablations, but this forgoes the efficiency
gains of top-\(K\) selection; an answer-string distant-supervision variant
(Appendix~\ref{app:controls}) suggests weaker signals can partially
substitute. Learning selection from weaker or self-supervised evidence remains
future work.

Second, our same-backbone experiments use Llama-3.2-1B and 3B;
we have not evaluated additional backbone families or whether the same
quality and efficiency margins over compression baselines transfer to
substantially larger models, where full-context
fine-tuning may become a stronger reference. Third, the extraction and injection
layers are chosen empirically. Although Appendix~\ref{app:layer_depth_ablation}
provides a diagnostic study, we do not propose a principled procedure for
selecting these layers on new backbone families. Our RULER \texttt{qa\_2}
evaluation extends to 16K contexts, where SeDeM remains stronger than ICAE but
its own quality declines as context length increases
(Section~\ref{sec:ruler_qa2}); we do not evaluate beyond 16K or on broader
long-context suites, where the trade-off between evidence coverage, selection
budget, and decompression cost may differ. In addition, our empirical evaluation is limited to long-context QA and
retrieval-oriented tasks; broad-coverage tasks such as summarization may
require different selection strategies and remain future work. Our evaluation relies primarily on lexical answer-quality metrics;
complementary semantic or human evaluation remains an important direction
for future work.

Fourth, with the 3B backbone, SeDeM remains below the full-context fine-tuned reference on MuSiQue. Our selection-budget
analysis (Appendix~\ref{app:k_sweep}) shows that \(K\) controls a
quality--context trade-off: larger budgets can recover additional evidence on
some datasets but process more positions above the injection layer, reducing the
efficiency margin. Adaptive or iterative selection that varies \(K\) per query
is a natural direction for future work.

Finally, our latency results follow a fully online protocol that includes
segmentation, encoding, compression, selection, and selective decompression.
Precomputing query-independent document memories could further improve
efficiency, but this system-level optimization is not evaluated here.

\section*{Potential Risks}
A common risk of LLM systems with aggressive long-context compression is
that relevant evidence may be omitted or distorted before generation. In
SeDeM, the decoder conditions on selected and reconstructed memory blocks
rather than the full context; if selection fails or relevant evidence is
omitted, the decoder may produce incomplete or unsupported answers.
Because the selector is trained from gold evidence annotations, any bias in
those annotations may propagate to which evidence the decoder sees.

SeDeM is also an efficiency method for long-context language-model
inference. By reducing the cost of processing long contexts, it could lower
the barrier to large-scale automated text generation, including low-quality
or misleading content. These risks are not unique to SeDeM, but they should
be considered when applying the method.

%% file: sections/appendix.tex
\section{Trainable Parameters}
\label{app:trainable_parameters}

The encoder is frozen throughout training, so the context-encoding pathway
remains independent of downstream task adaptation. The decoder backbone is
also frozen. We train the following components:
\begin{enumerate}
    \item \textbf{Compressor:}
    the input projection \(W_{\mathrm{in}}\) in
    Eq.~\ref{eq:input_projection}.

    \item \textbf{Decompressor:}
    the two-layer reconstruction network
    \(\{W_1^{\mathrm{dec}}, b_1^{\mathrm{dec}},
    W_2^{\mathrm{dec}}, b_2^{\mathrm{dec}}\}\)
    and its layer-normalization parameters in Eq.~\ref{eq:decompressor_mlp}.

    \item \textbf{Selection module:}
    the per-head projections
    \(\{W_Q^{(r)}, W_K^{(r)}\}_{r=1}^{R}\)
    in Eq.~\ref{eq:retrieval_similarity_method}.

    \item \textbf{Decoder adaptation:}
    LoRA factors added to the query, key, value, and output projection
    matrices of every decoder layer.
\end{enumerate}

For
\(X \in \{Q,K,V,O\}\),
the adapted decoder projection is
\begin{equation}
    W'_X
    =
    W_X
    +
    \frac{\alpha_{\mathrm{lora}}}{r_{\mathrm{lora}}}
    A_X B_X .
    \label{eq:lora_update}
\end{equation}
Here,
\(A_X \in \mathbb{R}^{d_{\mathrm{dec}} \times r_{\mathrm{lora}}}\)
and
\(B_X \in \mathbb{R}^{r_{\mathrm{lora}} \times d_{\mathrm{dec}}}\).
We use
\(r_{\mathrm{lora}}=64\)
and
\(\alpha_{\mathrm{lora}}=128\).
Only the LoRA factors are updated; the original decoder weights remain fixed.

\section{Two-Stage Optimization}
\label{app:two_stage_optimization}

Stage~1 trains the compressor and decompressor while the selector module and decoder LoRA adapters remain inactive. The objective combines context-window next-token reconstruction, decoder distillation from the raw-context teacher path, and hidden-state reconstruction \(\mathcal{L}_{\mathrm{rec}}\). The encoder and decoder pretrained weights remain frozen.

Stage~2 activates the full query-conditioned pipeline. The compressor and
decompressor are initialized from Stage~1, the selection module is randomly
initialized, and LoRA adapters are attached to the decoder. We update
\(W_{\mathrm{in}}\),
the decompressor parameters,
the selector projections
\(\{W_Q^{(r)}, W_K^{(r)}\}_{r=1}^{R}\),
and the LoRA factors
\(\{A_X,B_X\}_{X\in\{Q,K,V,O\}}\).
The encoder and decoder base weights remain frozen.

\section{Selection Supervision}
\label{app:retrieval_supervision}

The Top-\(K\) operator is discrete, so
\(\mathcal{L}_{\mathrm{LM}}\)
does not directly supervise the selection scores \(s_n\). We therefore use
block-level evidence labels
\(e_n \in \{0,1\}\).
Let
\[
    \mathcal{P} = \{n : e_n = 1\},
    \qquad
    \mathcal{N} = \{n : e_n = 0\}.
\]

The selection objective combines an InfoNCE term with a pairwise margin term:
\begin{equation}
    \mathcal{L}_{\mathrm{ret}}
    =
    \mathcal{L}_{\mathrm{InfoNCE}}
    +
    \lambda_{\mathrm{margin}}
    \mathcal{L}_{\mathrm{margin}}.
    \label{eq:loss_ret}
\end{equation}

The InfoNCE term encourages evidence-bearing blocks to score above the full
candidate set:
\begin{equation}
    \mathcal{L}_{\mathrm{InfoNCE}}
    =
    -
    \frac{1}{|\mathcal{P}|}
    \sum_{n^{+} \in \mathcal{P}}
    \log
    \frac{
        \exp(s_{n^{+}}/\tau)
    }{
        \sum_{n=1}^{S} \exp(s_n/\tau)
    }.
    \label{eq:loss_infonce}
\end{equation}

The margin term enforces a minimum score gap between positive and negative
blocks:
\begin{equation}
\mathcal{L}_{\mathrm{margin}}
=
\frac{1}{|\mathcal{P}|\,|\mathcal{N}|}
\sum_{n^{+}\in\mathcal{P}}\sum_{n^{-}\in\mathcal{N}}
\left[\gamma - s_{n^{+}} + s_{n^{-}}\right]_{+},
\label{eq:loss_margin}
\end{equation}
where \([x]_{+}=\max(0,x)\).

The InfoNCE term shapes the global block ranking, while the margin term
penalizes positive-negative pairs whose score separation is insufficient. The values of
\(\tau\),
\(\gamma\), and
\(\lambda_{\mathrm{margin}}\)
are reported in the experimental setup.

\section{Gradient Flow}
\label{app:gradient_flow}

Gradients from
\(\mathcal{L}_{\mathrm{LM}}\)
update the decompressor,
\(W_{\mathrm{in}}\),
and the decoder LoRA factors through the selected reconstructed states
\(\widehat{H}_{\mathcal{I}(Q)}\)
and the decoder continuation layers. Gradients from
\(\mathcal{L}_{\mathrm{ret}}\)
update the selection projections
\(\{W_Q^{(r)},W_K^{(r)}\}_{r=1}^{R}\)
and also propagate through the memory tokens \(B_n\) into
\(W_{\mathrm{in}}\).

Thus,
\(W_{\mathrm{in}}\)
is shaped by both selection and generation: it must produce memory tokens
that support evidence selection and downstream reconstruction.

\section{Inference}
\label{app:inference}

At inference time, all training losses are inactive. The model computes the selection scores \(s_n\), selects the top-\(K\) memory blocks, decompresses
the selected memories, injects the reconstructed hidden states at
\(\ell_{\mathrm{inject}}\), and generates the answer autoregressively.
Evidence labels are not required at inference.

\section{Additional Experimental Details}
\label{app:experimental_details}

\subsection{Dataset Processing}
\label{app:dataset_processing}

Table~\ref{tab:datasets} summarizes the tasks and context lengths of the four
benchmarks, and Table~\ref{tab:dataset_appendix_details} summarizes the
dataset sources, training/validation splits, and evidence-to-segment mapping
used for selection supervision.

\begin{table}[t]
\centering
\small
\setlength{\tabcolsep}{3pt}
\renewcommand{\arraystretch}{0.95}
\begin{tabular}{@{}lccc@{}}
\toprule
Dataset & Task & Avg. len. & Max len. \\
\midrule
HotpotQA-Dist. & 2-hop QA & 1{,}299 & 3{,}711 \\
2WikiMHQA & multi-hop QA & 834 & 7{,}000 \\
MuSiQue & 2--4-hop QA & 2{,}288 & 6{,}026 \\
QASPER & long-doc QA & 5{,}248 & 34{,}768 \\
\bottomrule
\end{tabular}
\caption{Long-context QA datasets. Lengths in Llama-3 tokens.}
\label{tab:datasets}
\end{table}

For HotpotQA-Distractor, we use the 10-paragraph distractor split from
\texttt{hotpotqa/hotpot\_qa}, configuration \texttt{distractor}. Each example
contains two gold supporting paragraphs and eight distractors. The gold
supporting-paragraph identifiers are mapped to positive segments.

For 2WikiMultiHopQA, we use \texttt{xanhho/2WikiMultihopQA}. For MuSiQue, we
use \texttt{dgslibisey/MuSiQue}. Both datasets provide paragraph-level
candidate contexts with gold supporting-paragraph labels, which we map to
positive selection segments.

For QASPER, we use \texttt{allenai/qasper}. The paper text is divided into contiguous fixed-length segments. We use the annotator-provided \texttt{evidence} field as selection supervision: a
QASPER segment is marked positive if it overlaps at least one evidence span.

All datasets are used for research training and evaluation in the same task
family for which they were released; we do not redistribute modified dataset
copies.

\begin{table*}[t]
\centering
\small
\setlength{\tabcolsep}{3pt}
\renewcommand{\arraystretch}{0.95}
\begin{tabular}{@{}l p{2.3cm} p{4.0cm} p{6.2cm}@{}}
\toprule
Dataset & Train / Val & Source & Evidence-to-segment mapping \\
\midrule

HotpotQA-Dist.
& 90{,}447 / 7{,}405
& \texttt{hotpotqa/hotpot\_qa}, \texttt{distractor}
& Gold supporting-paragraph identifiers are mapped to the corresponding positive selection segments. \\

2WikiMHQA
& 167{,}454 / 12{,}576
& \texttt{xanhho/2WikiMultihopQA}
& Gold supporting-paragraph labels are mapped to the corresponding positive segments. \\

MuSiQue
& 19{,}938 / 2{,}417
& \texttt{dgslibisey/MuSiQue}
& Gold supporting-paragraph labels are mapped to the corresponding positive segments. \\

QASPER
& 2{,}322 / 945
& \texttt{allenai/qasper}
& A segment is marked positive if it overlaps at least one annotator-provided evidence span. \\

\bottomrule
\end{tabular}
\caption{
Dataset sources, splits, and evidence-label mapping. Evidence labels are used only to supervise the selection loss during training and are not available at inference.
}
\label{tab:dataset_appendix_details}
\end{table*}

\subsection{Training Configuration}
\label{app:training_configuration}

Objective functions are defined in Section~\ref{sec:training_objective}.
This section reports concrete optimization settings.

\paragraph{Stage 1: reconstruction pretraining.}
The compressor and decompressor are trained on 300M tokens of SlimPajama
(\texttt{DKYoon/SlimPajama-6B}) for five epochs using the Stage~1 objective in Eq.~\ref{eq:loss_stage1}. This objective combines context-window next-token reconstruction, decoder distillation, and the hidden-state reconstruction loss \(\mathcal{L}_{\mathrm{rec}}\). The selection module and LoRA adapters are inactive in this stage.

\paragraph{Stage 2: selection-supervised task training.}
We initialize the compressor and decompressor from Stage~1, attach LoRA
adapters to the decoder, randomly initialize the selection module, and train
under the combined objective in Eq.~\ref{eq:loss_total} for up to three epochs
per dataset. For the selection objective in Eq.~\ref{eq:loss_ret}, we use
temperature \(\tau=0.07\), margin \(\gamma=2.0\), and
\(\lambda_{\mathrm{margin}}=0.5\). The overall loss weights are
\(\lambda_{\mathrm{ret}}=1.0\) and
\(\lambda_{\mathrm{rec}}=0.1\).

We use AdamW with separate learning rates:
\(1{\times}10^{-4}\) for the compressor and LoRA factors, and
\(5{\times}10^{-4}\) for the selection projections
\(\{W_Q^{(r)},W_K^{(r)}\}_{r=1}^{R}\).
Precision and gradient clipping match Stage~1.

\paragraph{LoRA configuration.}
We adapt the frozen decoder with LoRA at rank
\(r_{\mathrm{lora}}=64\)
and scaling
\(\alpha_{\mathrm{lora}}=128\).
Adapters are inserted into all four attention projections
\(\{W_Q,W_K,W_V,W_O\}\)
in every decoder layer.

\section{Implementation Details}
\label{app:implementation_details}

Our method is implemented in PyTorch with HuggingFace Transformers. All runs
use bfloat16 mixed precision and FlashAttention-2 or SDPA attention where
available. Generation is greedy with a 64-token cap for
HotpotQA-Distractor, 2WikiMultiHopQA, and MuSiQue, and a 128-token cap for
QASPER; decoding terminates earlier when EOS is emitted.

\paragraph{Online timing protocol.}
All reported TTFT values follow a fully online protocol. For our method,
timing begins when both the context and query are provided and includes:
context segmentation, encoder extraction at
\(\ell_{\mathrm{extract}}\), compression, memory-bank assembly, query
encoding for selection, selection scoring, decompression of the selected
memory blocks, decoder prefix processing up to
\(\ell_{\mathrm{inject}}\), hidden-state injection, continuation through the
remaining decoder layers, and emission of the first answer token. We do not
cache or precompute document-side memories for TTFT reporting.

The batched compressor is bit-exact to the sequential implementation because
all compressor operations are applied independently per segment or per feature.
The early-exit backbone forward is also bit-exact because it reuses the
original backbone computation up to layer \(\ell_{\mathrm{extract}}\) and skips only layers whose outputs are not consumed by the compressor or selector. These optimizations affect latency but not EM, F1, or ROUGE-L.

\section{Reproduction Details for Baselines}
\label{app:baseline_reproduction}

We evaluate LongLLMLingua~\citep{jiang2024longllmlingua}, ICAE~\citep{ge2024icae},
HMT~\citep{he2025hmt}, Activation Beacon~\citep{zhang2024activationbeacon}, and 500xCompressor (500x;
\citealp{li2025generalized}) under the same downstream train/validation splits
used for our method. The reproduced trainable baselines are adapted to
\texttt{meta-llama/Llama-3.2-1B} and \texttt{meta-llama/Llama-3.2-3B}, and
evaluated on HotpotQA-Distractor, 2WikiMultiHopQA, MuSiQue, and QASPER. To
isolate differences in the compression mechanism rather than dataset
preprocessing, all methods use the same source splits and answer-evaluation
pipeline. Trainable baselines use the same instruction-formatted training and
validation files where applicable; method-specific preprocessing constraints
are described below.

\paragraph{Shared implementation environment.}
The baseline reproduction experiments are run with PyTorch 2.5.1 and ROCm 6.2 on AMD MI300X/MI325X/MI250X/MI210 GPUs. We use Python 3.9 with pinned versions of \texttt{transformers==4.46.3}, \texttt{peft==0.11.1}, \texttt{accelerate==1.10.1}, and \texttt{deepspeed==0.14.4}. Training uses bf16 mixed precision throughout. Since the original baseline codebases were developed for different model families and software versions, a small number of compatibility patches were necessary: AB was extended to support the Llama-3 RoPE configuration schema, and 500x was updated to remove a hard-coded hidden size and migrate from the deprecated tuple-style \texttt{past\_key\_values} interface to \texttt{DynamicCache}. These are implementation compatibility changes only and do not alter the baseline algorithms. We also replace DeepSpeed \texttt{FusedAdam} with \texttt{torch.optim.AdamW} because the available fused optimizer build is incompatible with the AMD cluster environment.

\paragraph{Pretraining corpora.}
Two upstream corpora used in the original works were not directly available in our environment. For AB, we substitute \texttt{DKYoon/SlimPajama-6B} for RedPajama-1T-Sample while preserving the same 200M-token continual-pretraining budget. For 500x, we substitute \texttt{UniverseTBD/arxiv-abstracts-large} for the Kaggle arXiv dataset used by the original implementation.

\subsection{LongLLMLingua}
\label{app:longllmlingua_details}

We evaluate LongLLMLingua~\citep{jiang2024longllmlingua} as a training-free
hard prompt-compression baseline. Unlike the trainable soft-memory baselines,
LongLLMLingua does not use supervised fine-tuning, LoRA adapters, or
task-specific parameter updates in our setup. For each example, it compresses
the raw context and then passes the compressed prompt together with the
question to the same base LLM used for generation. We use the corresponding
Llama-3.2-Base backbone as both the scorer model and the generator model, with
bfloat16 inference, greedy decoding, and no chat template. This is a stricter
setting than the original LongLLMLingua evaluation, which primarily used
stronger instruction/chat models as target LLMs.

We use a fixed 4$\times$ nominal compression setting, implemented with
compression rate 0.25. The LongLLMLingua configuration uses question-conditioned
compression with \texttt{condition\_compare=True},
\texttt{condition\_in\_question=after},
\texttt{rank\_method=longllmlingua}, \texttt{reorder\_context=sort},
\texttt{dynamic\_context\_compression\_ratio=0.4},
\texttt{context\_budget=+100}, and no sentence-level filtering. Unless
otherwise stated, we report the strict zero-shot setting with no demonstration
examples.

Because LongLLMLingua keeps selected raw tokens verbatim, its compression
interface differs from continuous-memory baselines such as ICAE,
500xCompressor, and SeDeM. The comparison should therefore be
interpreted as selective token retention versus lossy continuous-memory
compression under a similar nominal compression budget, rather than as an
identical latent-memory budget. We use the same answer-normalization and
evaluation pipeline as SeDeM, including lowercasing, punctuation
removal, article removal, max-over-reference scoring, token-level F1, EM, and
ROUGE-L. For compatibility with recent Transformers cache objects, we patch the
LongLLMLingua perplexity-scoring path to convert \texttt{DynamicCache} objects
to the list-of-key-value format expected by the released implementation.

\subsection{ICAE}
\label{app:icae_details}

We reproduce ICAE~\citep{ge2024icae} as a soft-memory compression baseline
using Llama-3.2-\{1B,3B\}-Base backbones. Following the ICAE formulation, the
model uses learned memory tokens to encode the context into continuous memory
states that are consumed directly by a frozen decoder. Our implementation uses
two Llama-3.2 instances initialized from the same checkpoint: an encoder with
LoRA adapters on the query and value projections, and a fully frozen decoder
used in evaluation mode. The trainable parameters are the encoder LoRA adapters
and a separate memory-token embedding table. The decoder backbone is not
updated.

ICAE is trained with a two-stage recipe. In Stage~1, we train on SlimPajama
using autoencoding and language-modeling objectives: the autoencoding loss
reconstructs the context from memory states, and the language-modeling loss
predicts the continuation from those states. We use memory size 128, context
length 512, next-token continuation length 128, LoRA rank 128, dropout 0.05,
bfloat16 training, AdamW with learning rate \(1{\times}10^{-4}\), cosine decay,
warmup ratio 0.05, and gradient clipping at 2.0. For the 1B backbone we
pretrain on 1B tokens; for the 3B backbone we pretrain on 300M tokens.

In Stage~2, we continue training the same trainable parameters on the
downstream QA data using the ICAE QA decoding mode. The decoder remains frozen,
while the encoder LoRA adapters and memory-token embeddings are updated. We use
AdamW with learning rate \(5{\times}10^{-5}\), cosine decay, warmup ratio 0.05,
and gradient clipping at 2.0. Downstream training transforms are shared with
SeDeM so that the training examples are matched where applicable;
evaluation uses the full distractor context for distractor-style datasets.
Generation is greedy, with the same answer length limits and the same
normalized EM, token-level F1, and ROUGE-L evaluation as SeDeM. When
multiple downstream checkpoints are saved, we select the checkpoint with the
lowest validation loss, following the shared checkpoint-selection rule.

Our ICAE reproduction differs from the original ICAE setup in three main
respects: we use Llama-3.2-Base backbones rather than the original model
family, we use explicit Stage~1 token budgets matched to our compute setting,
and we reuse the same downstream preprocessing and answer-normalization
pipeline as SeDeM to ensure a controlled comparison.

\subsection{HMT}
\label{app:hmt_details}

We include HMT as a hierarchical memory-based baseline for long-context
processing. We adapt the HMT training and evaluation pipeline to the same
Llama-3.2-\{1B,3B\} backbones and downstream QA files used by the other
baselines. The model is trained separately for each downstream dataset and
evaluated with the same answer-generation limits and token-level F1/ROUGE-L
metrics used for all methods.

For inference-time parameter accounting, we count only modules that remain
active at inference. HMT's instantiated \texttt{MemoryMap} module is excluded
from the inference-active parameter count because it is not used during
inference in our reproduction.

\subsection{Activation-Beacon}
\label{app:ab_details}

We follow the two-stage training recipe of Activation-Beacon: continual pretraining to introduce beacon compression, followed by task-specific supervised fine-tuning.

\paragraph{Stage 1: continual pretraining.}
We initialise from \texttt{Llama-3.2-\{1B,3B\}} and train for one epoch on a 200M-token SlimPajama-6B subsample. Sequence lengths are grouped in the range 2{,}400--20{,}000 tokens. The beacon window size and stride are both set to 1{,}024 tokens. At each step, the compression ratio is sampled from $\{2,4,8,16,32\}$. We use the full-coverage prefix-style attention configuration with \texttt{beacon\_attend\_prev=True} and \texttt{beacon\_sink\_size=1}. Beacon states are produced through the $Q/K/V$ projections and interleaved with the token sequence, matching the official implementation. Attention is computed using SDPA. Optimisation uses DeepSpeed ZeRO-2 with AdamW, per-device batch size 1, and gradient accumulation 8.

\paragraph{Stage 2: task-specific SFT.}
For each downstream task, we initialise from the Stage-1 checkpoint and train for one epoch on the corresponding QA training split. The maximum sequence length is set to 16{,}384 tokens. The beacon configuration is preserved, except that the compression-ratio set is narrowed to $\{2,4,8\}$ during SFT. We use learning rate $1\times10^{-5}$ with a linear schedule, the Llama-3 chat template, and the same effective batch size of 8.

\paragraph{Deviations from the original AB setup.}
Our AB reproduction differs from \citet{zhang2024activationbeacon} in four respects: (i) SlimPajama-6B replaces RedPajama-1T-Sample under the same 200M-token budget; (ii) AdamW replaces DeepSpeed FusedAdam for hardware compatibility; (iii) HotpotQA is evaluated in the distractor setting rather than a single-context formulation; and (iv) the same configuration is applied to both 1B and 3B backbones, whereas the original study focuses on larger models.

The reported evaluation explicitly passes the intended \texttt{eos\_token\_id}, preventing generation from continuing beyond the predicted answer. Training, supervised fine-tuning, and the saved checkpoints are unchanged.

\paragraph{Reproduction sensitivity.}
During our reproduction audit, we found Activation Beacon to be sensitive
to several implementation and data-format choices. In particular,
generation termination depended on the supervised target/EOS convention
and chat-template format. With the Llama-3.2-Base backbone, some
chat-formatted configurations systematically continued generation to the
maximum token limit, whereas target formats aligned with the backbone's
EOS behavior terminated normally. We therefore explicitly pass the
intended \texttt{eos\_token\_id} during evaluation.

We also found that the beacon window must be interpreted relative to the
actual downstream context length. Examples shorter than the configured
1,024-token beacon window can contain no beacon positions and therefore
do not exercise the intended beacon-compression mechanism. In our
2WikiMultiHopQA evaluation, this occurred for approximately 61\% of
examples. 

\subsection{500xCompressor}
\label{app:500x_details}

We reproduce 500xCompressor using the official two-stage pipeline: reconstruction-oriented pretraining followed by task-specific QA fine-tuning. The model compresses up to \texttt{max\_length}=500 context tokens into \texttt{num\_mem}=4 learned memory tokens, corresponding to a 125$\times$ per-segment compression ratio. We choose this configuration to keep the per-segment latent-memory budget comparable to the other memory-based baselines in our study.

\paragraph{Stage 1: reconstruction pretraining.}
We initialise from \texttt{Llama-3.2-\{1B,3B\}} and train for 3 epochs on \texttt{UniverseTBD/arxiv-abstracts-large} using the reconstruction objective from the original 500x implementation, where the decoder reconstructs the input passage from the memory tokens. The LoRA configuration is $r=64$, $\alpha=32$, and dropout $0.05$, applied to the default PEFT causal-LM attention targets (\texttt{q\_proj} and \texttt{v\_proj}). Optimisation uses DeepSpeed ZeRO-3 with AdamW, learning rate $1\times10^{-4}$, a constant-with-warmup schedule with 300 warmup steps, and per-device batch size 8. After pretraining, LoRA weights are consolidated from the DeepSpeed shards for downstream loading.

\paragraph{Stage 2: task-specific SFT.}
For each task, we load the Stage-1 LoRA weights onto the corresponding Llama-3.2 backbone and continue supervised fine-tuning for up to 10 epochs. We select the checkpoint with the lowest validation loss. LoRA hyperparameters are unchanged. We reduce the learning rate to $5\times10^{-5}$ with 200 warmup steps and use per-device batch size 8. The question-answer budget is set to \texttt{max\_qa\_len}=64; when necessary, the question is truncated first to ensure that the answer retains at least 8 tokens.

\paragraph{Global context truncation caveat.}
A key implementation constraint of 500x is that the 500-token context limit is applied globally during both SFT and inference. This is consequential for interpreting downstream results. On HotpotQA-Distractor and 2WikiMultihopQA, relevant evidence often appears sufficiently early in the concatenated context that a 500-token prefix can still retain useful supervision. In contrast, MuSiQue contexts are longer and gold evidence is more dispersed, so relevant passages are frequently excluded by the global truncation window. For QASPER, we pre-chunk the document context into 500-token windows during data preparation, so this particular truncation issue does not arise in the same way.

\paragraph{Deviations from the original 500x setup.}
Our 500x reproduction differs from \citet{li2025generalized} in three respects: (i) \texttt{UniverseTBD/arxiv-abstracts-large} replaces the Kaggle arXiv corpus used in the original codebase; (ii) we use \texttt{num\_mem}=4 and \texttt{max\_length}=500, yielding 125$\times$ compression, to align the memory budget with the other compressed-memory baselines; and (iii) we evaluate Llama-3.2-\{1B,3B\} rather than the LLaMA-3-8B-Instruct backbone considered in the original paper.

\section{Efficiency and Overhead Details}
\label{app:efficiency_overhead}
\label{app:effective_context_length}

\paragraph{Per-dataset efficiency compared with ICAE.}
Table~\ref{tab:efficiency_varb_icae} reports per-dataset online TTFT and
decoding throughput for SeDeM and ICAE, summarized in
Section~\ref{sec:inference_efficiency}.

\begin{table*}[t]
\centering
\small
\setlength{\tabcolsep}{3pt}
\renewcommand{\arraystretch}{0.95}
\begin{tabular}{@{}l cc cc@{}}
\toprule
& \multicolumn{2}{c}{\textit{Llama-3.2-1B-Base}}
& \multicolumn{2}{c}{\textit{Llama-3.2-3B-Base}} \\
\cmidrule(lr){2-3} \cmidrule(lr){4-5}
Dataset
& TTFT Ours/ICAE $\downarrow$
& Tok/s Ours/ICAE $\uparrow$
& TTFT Ours/ICAE $\downarrow$
& Tok/s Ours/ICAE $\uparrow$ \\
\midrule
Hotpot
& 47.43/58.52 \textcolor{green!45!black}{(1.23$\times$)}
& 77.41/71.87 \textcolor{green!45!black}{(1.08$\times$)}
& 65.43/125.97 \textcolor{green!45!black}{(1.93$\times$)}
& 46.51/41.76 \textcolor{green!45!black}{(1.11$\times$)} \\

MuSiQue
& 62.63/79.28 \textcolor{green!45!black}{(1.27$\times$)}
& 77.43/72.41 \textcolor{green!45!black}{(1.07$\times$)}
& 91.34/193.96 \textcolor{green!45!black}{(2.12$\times$)}
& 45.30/40.95 \textcolor{green!45!black}{(1.11$\times$)} \\

2Wiki
& 33.49/43.14 \textcolor{green!45!black}{(1.29$\times$)}
& 77.98/72.53 \textcolor{green!45!black}{(1.08$\times$)}
& 65.14/98.99 \textcolor{green!45!black}{(1.52$\times$)}
& 45.89/41.27 \textcolor{green!45!black}{(1.11$\times$)} \\

QASPER
& 42.70/143.34 \textcolor{green!45!black}{(3.36$\times$)}
& 78.29/70.49 \textcolor{green!45!black}{(1.11$\times$)}
& 96.50/364.10 \textcolor{green!45!black}{(3.77$\times$)}
& 42.95/40.69 \textcolor{green!45!black}{(1.06$\times$)} \\
\midrule
\textbf{Avg.}
& 46.56/81.07 \textcolor{green!45!black}{\textbf{(1.74$\times$)}}
& 77.78/71.82 \textcolor{green!45!black}{\textbf{(1.08$\times$)}}
& 79.60/195.76 \textcolor{green!45!black}{\textbf{(2.46$\times$)}}
& 45.16/41.17 \textcolor{green!45!black}{\textbf{(1.10$\times$)}} \\
\bottomrule
\end{tabular}
\caption{Efficiency compared with ICAE. TTFT is in milliseconds and
throughput is in tokens per second. All SeDeM and ICAE timing measurements
were obtained on an exclusively allocated NVIDIA A100-SXM4-40GB GPU with
batch size 1 and bfloat16 precision. Speedup is shown in parentheses.}
\label{tab:efficiency_varb_icae}
\end{table*}

\paragraph{ComprExIT TTFT comparison.}
Table~\ref{tab:comprexit_ttft} reports the online TTFT measurements for the
matched ComprExIT comparison in Section~\ref{sec:comprexit_comparison}.

\begin{table}[t]
\centering
\small
\setlength{\tabcolsep}{4pt}
\renewcommand{\arraystretch}{0.95}
\begin{tabular}{@{}lcc@{}}
\toprule
Configuration & Hotpot & 2Wiki \\
\midrule
SeDeM learned-\(K{=}2\) & 47.43 & 33.49 \\
SeDeM full-bank         & 50.01 & 49.81 \\
ComprExIT (reference)   & 111.7 & 110.7 \\
ComprExIT (matched)     & 373.8 & 377.9 \\
\bottomrule
\end{tabular}
\caption{
Online TTFT (ms) on HotpotQA-Distractor and 2WikiMHQA for the
ComprExIT comparison. Measured on an exclusively allocated
A100-SXM4-40GB with batch size 1, bfloat16, and context length
1,536; TTFT includes compression and decoder prefill.
}
\label{tab:comprexit_ttft}
\end{table}

\paragraph{Effective decoder-side context length.}

Table~\ref{tab:length_matched_controls} reports the decoder-side context length induced by different selection budgets. This table makes the quality--efficiency trade-off explicit: increasing \(K\) exposes more selected memory blocks to the decoder but also increases the number of positions processed above the injection layer.

\begin{table}[t]
\centering
\small
\setlength{\tabcolsep}{3.2pt}
\renewcommand{\arraystretch}{1.08}
\begin{tabular}{@{}lccccc@{}}
\toprule
Dataset 
& Blocks 
& $K_{\text{low}}$ 
& Len$_{\text{low}}$ 
& $K_{\text{high}}$ 
& Len$_{\text{high}}$ \\
\midrule
HotpotQA-Dist. 
& 10 
& 2 
& $\sim$331 
& 8 
& $\sim$1099 \\

2WikiMHQA 
& 10 
& 2 
& $\sim$331 
& 10 
& $\sim$1355 \\

MuSiQue 
& 20 
& 2 
& $\sim$331 
& 10 
& $\sim$1355 \\

QASPER 
& 64 
& 8 
& $\sim$1104 
& 32 
& $\sim$4176 \\
\bottomrule
\end{tabular}
\caption{
Effective decoder-side context lengths for representative selection-budget
settings. \(K_{\text{low}}\) and \(K_{\text{high}}\) are sweep values used to
illustrate the quality--efficiency trade-off; \(K_{\text{high}}\) is not the
maximum possible selection budget. Lengths are approximated as \(128K\) plus
query and prompt-formatting tokens.
}
\label{tab:length_matched_controls}
\end{table}

\paragraph{Cross-method inference-time parameter overhead.}
\label{app:inference_parameter_overhead}

Table~\ref{tab:inference_parameter_overhead} compares the additional
parameters that remain active at inference time across compression methods.
The frozen Llama-3.2 backbones contain 1.236B and 3.213B parameters for the
1B and 3B settings, respectively. For HMT, the instantiated
\texttt{MemoryMap} module is excluded because it is not used at inference:
8.39M parameters for the 1B model and 18.87M for the 3B model.

For our method, all inference-active modules are included: the compressor,
decompressor, selector, and LoRA adapters. The resulting total overhead is
47.22M parameters for Llama-3.2-1B and 112.25M for Llama-3.2-3B, corresponding
to \(3.82\%\) and \(3.49\%\) of the frozen backbones.

\begin{table}[t]
\centering
\small
\setlength{\tabcolsep}{3pt}
\renewcommand{\arraystretch}{0.95}
\begin{tabular}{@{}llr@{}}
\toprule
Method
& Added modules
& \begin{tabular}[c]{@{}c@{}}Added /\\ backb. (\%)\end{tabular} \\
\midrule

\multicolumn{3}{c}{\textit{Llama-3.2-1B-Base}} \\
\midrule
ICAE
& 0.54M mem. emb.
& 0.04 \\

HMT
& 0.020M mem. tok.
& 0.002 \\

500x
& 0.008M mem. tok.
& 0.001 \\

Act. Beacon
& 100.67M beacon QKV+emb.
& 8.15 \\

\textbf{Ours}
& 33.59M comp.+decomp.+sel.
& 2.72 \\

\midrule
\multicolumn{3}{c}{\textit{Llama-3.2-3B-Base}} \\
\midrule
ICAE
& 0.81M mem. emb.
& 0.03 \\

HMT
& 0.031M mem. tok.
& 0.001 \\

500x
& 0.012M mem. tok.
& \(<0.001\) \\

Act. Beacon
& 440.40M beacon QKV+emb.
& 13.71 \\

\textbf{Ours}
& 75.55M comp.+decomp.+sel.
& 2.35 \\
\bottomrule
\end{tabular}
\caption{
Method-specific inference-time parameter overhead, excluding LoRA adapters.
The frozen Llama-3.2 backbones contain 1.236B and 3.213B parameters for the
1B and 3B settings, respectively. For HMT, instantiated
\texttt{MemoryMap} parameters unused at inference are excluded.
For Ours, the reported modules include the compressor, decompressor, and
selector.
}
\label{tab:inference_parameter_overhead}
\end{table}

\paragraph{Parameter and FLOP breakdown for SeDeM.}
\label{app:inference_overhead_app}

We analyze inference-time overhead using additional trainable parameters and theoretical per-query FLOPs. The parameter count includes the compressor, decompressor, selector, and decoder LoRA adapters. The FLOP analysis estimates architecture-level online computation and is independent of hardware-specific effects. 

\paragraph{Parameter overhead.}
Table~\ref{tab:ours_inference_params} shows the parameter footprint of our memory components. They add \(47.22\)M parameters for Llama-3.2-1B and \(112.25\)M parameters for Llama-3.2-3B, corresponding to \(3.82\%\) and \(3.49\%\) of the frozen backbones.

\begin{table}[t]
\centering
\small
\setlength{\tabcolsep}{3pt}
\renewcommand{\arraystretch}{0.95}
\begin{tabular}{lcc}
\toprule
Module & Llama-3.2-1B & Llama-3.2-3B \\
\midrule
Compressor
& 4.21M
& 9.46M \\

Decompressor
& 20.99M
& 47.21M \\

Selector
& 8.40M
& 18.89M \\

LoRA
& 13.63M
& 36.70M \\
\midrule
Total
& 47.22M (3.82\%)
& 112.25M (3.49\%) \\
\bottomrule
\end{tabular}
\caption{
Inference-time parameter breakdown. Percentages are relative to the frozen Llama-3.2-1B and Llama-3.2-3B
backbones.
}
\label{tab:ours_inference_params}
\end{table}

\paragraph{Theoretical inference FLOPs.}

Table~\ref{tab:inference_flops_query} compares per-query inference FLOPs at \(N_{\mathrm{ctx}}=2048\), \(N_q=30\), \(N_{\mathrm{ans}}=16\). SeDeM reduces total FLOPs because encoding stops at \(\ell_{\mathrm{extract}}\) and the decoder processes only the \(K\) reconstructed blocks above \(\ell_{\mathrm{inject}}\); ICAE and Activation Beacon require full-depth processing, and HMT pays for segmented backbone passes. The 500xCompressor estimate is not directly comparable, since its recipe processes only a single 500-token chunk.

\begin{table}[t]
\centering
\small
\setlength{\tabcolsep}{3pt}
\renewcommand{\arraystretch}{0.95}
\begin{tabular}{@{}lrrr@{}}
\toprule
Method
& Encode
& Decode
& Total \\
\midrule

\multicolumn{4}{c}{\textit{Llama-3.2-1B}} \\
\midrule
ICAE
& 4219.8
& 868.6
& 5088.4 \\

HMT
& 3418.0
& 117.7
& 3535.7 \\

500xCompressor\(^{*}\)
& 1014.2
& 106.0
& 1120.1 \\

Activation Beacon
& 4219.8
& 870.7
& 5090.5 \\

\textbf{Ours}
& 2892.9
& 350.7
& \textbf{3243.6} \\

\midrule
\multicolumn{4}{c}{\textit{Llama-3.2-3B}} \\
\midrule
ICAE
& 12091.7
& 2497.9
& 14589.6 \\

HMT
& 9866.4
& 329.2
& 10195.5 \\

500xCompressor\(^{*}\)
& 2928.5
& 295.1
& 3223.7 \\

Activation Beacon
& 12091.7
& 2503.9
& 14595.5 \\

\textbf{Ours}
& 4750.3
& 1003.0
& \textbf{5753.3} \\

\bottomrule
\end{tabular}
\caption{
Inference FLOPs per query. Values are reported in GFLOPs. \(^{*}\)The 500xCompressor estimate corresponds to the single-chunk
inference, which processes only a 500-token context chunk.
}
\label{tab:inference_flops_query}
\end{table}

\section{Effective-Rank Definition}
\label{app:erank_definition}

\begin{table}[h]
\centering
\small
\setlength{\tabcolsep}{3pt}
\renewcommand{\arraystretch}{0.95}
\begin{tabular}{lcc}
\toprule
Dataset & ICAE & SeDeM \\
\midrule
HotpotQA-Dist. & 53.7 & \textbf{80.2} \\
QASPER         & 54.1 & \textbf{92.3} \\
MuSiQue        & 60.6 & \textbf{74.7} \\
2Wiki          & 40.6 & \textbf{67.4} \\
\midrule
Average        & 52.3 & \textbf{78.7} \\
\bottomrule
\end{tabular}
\caption{Normalized effective rank of memory slots. Higher values indicate broader use of the available slot-rank capacity.}
\label{tab:normalized_erank}
\end{table}

We use normalized effective rank to measure whether compressed memory slots
collapse into redundant representations. Given a memory-slot matrix
\(\mathbf{M}\in\mathbb{R}^{N_M\times d}\), let
\(\{\sigma_i\}_{i=1}^{r}\) be its singular values, where
\(r=\min(N_M,d)\). We normalize the spectrum as
\[
    p_i = \frac{\sigma_i}{\sum_{j=1}^{r}\sigma_j},
\]
and define
\[
    \operatorname{eRank}(\mathbf{M})
    =
    \exp\left(
    -\sum_{i=1}^{r} p_i \log p_i
    \right).
\]
Because different methods may use different memory-slot budgets, we report
\(\operatorname{eRank}(\mathbf{M})/N_M\). Higher values indicate broader use
of the available slot-rank capacity.

\section{Learnable Compression without Block Structure}
\label{app:ablation_query_compressor}
\label{app:learnable_query_compressor}

We compare our block-structured compressor with a learnable-query compressor at
the same compression ratio. Our default compressor ties each memory slot to a
contiguous local chunk of the segment. The alternative uses a fixed set of
learnable latent queries that attend over the full segment, allowing each slot
to aggregate information from arbitrary token positions. This comparison tests
whether flexible global aggregation is preferable to block-structured local
compression for memory construction.

For a segment \(s\) with hidden states \(H_s \in \mathbb{R}^{T \times D}\), we
initialize \(M\) learnable query vectors
\(Q_0 \in \mathbb{R}^{M \times D}\), where \(M\) matches the number of memory
tokens used by our default compressor. Thus, the compression ratio is unchanged.
These queries are shared across examples and do not carry recurrent state
across segments.

For attention head \(h\), we compute
\begin{align}
Q^{h} &= \mathrm{LN}_{Q}(Q_0) W_Q^{h}, \nonumber \\
K_s^{h} &= \mathrm{LN}_{KV}(H_s) W_K^{h}, \nonumber \\
V_s^{h} &= \mathrm{LN}_{KV}(H_s) W_V^{h}.
\end{align}
The initial query-to-token attention is normalized over segment tokens:
\[
A^{\mathrm{init},h}
=
\operatorname{softmax}_{t}
\left(
\frac{Q^{h}(K_s^{h})^\top}{\sqrt{d_h}}
\right)
\in \mathbb{R}^{M \times T},
\]
where \(d_h\) is the per-head dimension.

To discourage multiple memory slots from repeatedly aggregating the same tokens,
we apply slot competition by renormalizing each token's attention mass across
memory slots:
\[
A_{i,t}^{h}
=
\frac{
A_{i,t}^{\mathrm{init},h}
}{
\sum_{j=1}^{M} A_{j,t}^{\mathrm{init},h} + \epsilon
},
\]
where \(\epsilon\) is a small constant for numerical stability. Each head
aggregates segment values as
\[
Z_s^{h} = A_s^{h} V_s^{h}.
\]

Inspired by Perceiver-style learned-query cross-attention~\citep{jaegle2021perceiverio},
we remove the output projection in this variant and add a residual connection
from the learnable queries. The \(i\)-th memory token is therefore
\[
m_{s,i}
=
Q_{0,i}
+
\mathrm{Concat}_{h}\left(Z_{s,i}^{h}\right).
\]

This variant is a stronger non-local compressor: it uses multi-head
query-to-token attention, slot competition to reduce duplicated token
assignment across memory slots, and a residual connection from the learned
queries. We evaluate it with the same learned top-\(K\) selector on
HotpotQA-Distractor.

\begin{table}[t]
\centering
\small
\setlength{\tabcolsep}{3pt}
\renewcommand{\arraystretch}{0.95}
\begin{tabular}{@{}lcc@{}}
\toprule
Selected blocks & Token-F1 & ROUGE-L \\
\midrule
\(K=2\)  & 33.2 & 33.1 \\
\(K=4\)  & 40.2 & 40.0 \\
\(K=8\)  & 51.6 & 51.3 \\
ALL      & \textbf{58.4} & \textbf{58.2} \\
\bottomrule
\end{tabular}
\caption{
HotpotQA-Distractor results for the learnable-query compressor on 1{,}000
validation examples. Scores are reported on a 0--100 scale. Accuracy increases
with the number of selected blocks and is highest when all blocks are used.
}
\label{tab:query-ksweep}
\end{table}

Table~\ref{tab:query-ksweep} shows that performance improves as more blocks are
retained and is highest when all ten blocks are used. This suggests that the
learnable-query compressor is less compatible with top-\(K\) selection: because
each slot can attend globally, information becomes less localized and harder to
associate with a specific selectable block.

\section{Extraction and Injection Depth}
\label{app:layer_depth_ablation}
This appendix reports a diagnostic layer-depth study rather than a full benchmark comparison. All experiments use a controlled HotpotQA-Distractor subset with 3{,}000 training examples and 1{,}000 validation examples, so the scores are not directly comparable to the full-dataset results in Table~\ref{tab:main_qa_results}. These runs disable selection and feed all segment reconstructions, isolating the effect of layer choice from selection quality.

\paragraph{Single-layer extraction and injection.}
Table~\ref{tab:layer_extraction_injection} evaluates single-layer extraction and injection choices. Across the tested settings, injection at layer 10 is consistently strongest, while extraction depth has a smaller effect. The best single-layer configuration extracts from \(\ell_{\mathrm{extract}}=16\) and injects at \(\ell_{\mathrm{inject}}=10\).

\begin{table}[t]
\centering
\small
\setlength{\tabcolsep}{3pt}
\renewcommand{\arraystretch}{0.95}
\caption{
Effect of single-layer extraction and decoder injection depth.
}
\label{tab:layer_extraction_injection}
\begin{tabular}{cccc}
\toprule
\(\ell_{\mathrm{extract}}\)
& \(\ell_{\mathrm{inject}}\)
& ROUGE-L
& F1 \\
\midrule
16 & 10 & \textbf{61.71} & \textbf{61.91} \\
12 & 10 & 61.68 & 61.90 \\
14 & 10 & 61.61 & 61.82 \\
10 & 10 & 60.97 & 61.08 \\
\midrule
12 & 12 & 59.79 & 60.00 \\
\midrule
14 & 14 & 57.60 & 57.93 \\
10 & 14 & 55.50 & 55.67 \\
\midrule
14 & 16 & 53.12 & 53.28 \\
16 & 16 & 48.88 & 49.04 \\
\bottomrule
\end{tabular}
\end{table}

\paragraph{Mixing extraction layers.}
Table~\ref{tab:mixed_extraction_layers} evaluates whether combining extraction layers improves the compressed representation. The \(\{10,12,14,16\}\) mixture improves over the best single-layer setting, while learned and uniform mixtures perform almost identically. This suggests that the gain comes mainly from using multiple intermediate layers, not from learning a highly specialized weighting. The wider \(\{4,10,16,22\}\) mixture performs worse, suggesting that distant layers are harder to combine into a decoder-usable representation.

\begin{table}[t]
\centering
\small
\setlength{\tabcolsep}{3pt}
\renewcommand{\arraystretch}{0.95}
\caption{
Effect of mixing encoder extraction layers with
\(\ell_{\mathrm{inject}}=10\) in the selector-disabled HotpotQA-Distractor
subset setting.
}
\label{tab:mixed_extraction_layers}
\begin{tabular}{lccc}
\toprule
Extraction layers
& Mixing
& ROUGE-L
& F1 \\
\midrule
10, 12, 14, 16
& learned
& \textbf{63.23}
& \textbf{63.51} \\

10, 12, 14, 16
& uniform
& 63.23
& 63.50 \\

4, 10, 16, 22
& learned
& 62.48
& 62.68 \\
\bottomrule
\end{tabular}
\end{table}

For the learned mixture over layers \(\{10,12,14,16\}\), the normalized weights are \(0.214, 0.144, 0.373,\) and \(0.269\), respectively.

\section{A Stability View of Layer Injection}
\label{app:stability_layer_injection}

This appendix provides a stability view of the layer-injection argument used to
motivate decoder-compatible hidden-state decompression. The argument is intended
only as architectural motivation for decoder-layer alignment. It does not prove that intermediate-layer injection is always better than input-level memory
tokens, and it does not imply exact recovery of full-context hidden states. In the implementation, \(\mathcal{L}_{\mathrm{rec}}\) aligns decompressed states to
projected encoder states, while decoder compatibility is further encouraged by
the context-window next-token loss, distillation loss, and Stage~2 QA loss.

Let \(P_\ell=\mathcal{D}(H^{(\ell)})\) denote the distribution of hidden states at decoder layer \(\ell\) when the decoder processes the selected context directly, and let \(\widehat P_\ell=\mathcal{D}(\widehat H^{(\ell)})\) denote the distribution of decompressed states injected at the same layer. In an idealized decoder-state reconstruction setting, paired examples \((H^{(\ell)},\widehat H^{(\ell)})\) form a valid coupling. Therefore,
\[
    \mathcal{W}_2^2(\widehat P_\ell,P_\ell)
    \leq
    \mathbb{E}
    \left[
        \left\|
            \widehat H^{(\ell)} - H^{(\ell)}
        \right\|_2^2
    \right].
\]
This follows directly from the definition of the Wasserstein distance, since \(\mathcal{W}_2\) is the infimum over all couplings and the paired reconstruction examples define one valid coupling.

Now let \(F_{\ell:L}\) denote the remaining decoder computation from layer \(\ell\) to the final layer \(L\). Suppose that \(F_{\ell:L}\) is locally \(K_{\ell:L}\)-Lipschitz on a neighborhood of the layer-\(\ell\) activation manifold that contains both \(H^{(\ell)}\) and \(\widehat H^{(\ell)}\) with high probability. This is a strong assumption for Transformer layers in general, so the bound should be interpreted only in the local regime where the injected states remain close to the decoder-layer activation manifold.

Let \(Q_\ell=(F_{\ell:L})_{\#}P_\ell\) and \(\widehat Q_\ell=(F_{\ell:L})_{\#}\widehat P_\ell\) be the pushforward distributions after the remaining decoder layers. By the standard Lipschitz pushforward inequality for Wasserstein distance,
\[
    \mathcal{W}_2(\widehat Q_\ell,Q_\ell)
    \leq
    K_{\ell:L}\,
    \mathcal{W}_2(\widehat P_\ell,P_\ell).
\]
Combining this with the reconstruction coupling bound gives
\[
    \mathcal{W}_2(\widehat Q_\ell,Q_\ell)
    \leq
    K_{\ell:L}
    \left(
        \mathbb{E}
        \left[
            \left\|
                \widehat H^{(\ell)} - H^{(\ell)}
            \right\|_2^2
        \right]
    \right)^{1/2}.
\]
Thus, if the decompressed states are close to the decoder-layer states they are meant to approximate, then the discrepancy after the remaining decoder layers is controlled up to the local stability constant \(K_{\ell:L}\).

This view also explains why input-level memory tokens are not directly comparable to intermediate-layer injection. Input-level methods introduce synthetic states near the bottom of the decoder and rely on lower layers to transform them into useful internal representations. In contrast, layer injection places decompressed states at the layer where they are consumed. This avoids requiring the memory representation to pass through the full lower decoder stack before becoming useful to later layers. We use this only as architectural intuition: the mismatch at the input layer and the mismatch at layer \(\ell\) are not directly comparable, and Lipschitz products across many Transformer layers can be loose or even misleading.

\section{Effect of Selection Budget \(K\)}
\label{app:k_sweep}

We evaluate how the selection budget \(K\) affects answer quality on HotpotQA-Distractor and 2WikiMultiHopQA. Table~\ref{tab:k_sweep} reports exact match (EM), F1, and ROUGE-L for different values of \(K\). For HotpotQA-Distractor, the evaluation uses the 1B (+ LoRA) model on the full validation set of 7,405 examples. For 2WikiMultiHopQA, we use the 1B full checkpoint.

The two datasets show different behavior. On HotpotQA-Distractor, increasing \(K\) does not improve answer quality: performance is best at \(K=2\), remains similar at \(K=4\), and then gradually decreases. This suggests that larger selection budgets can introduce additional weakly useful context into the reconstructed memory. On 2WikiMultiHopQA, score-based selection improves as \(K\) increases up to \(K=8\), after which performance saturates. This suggests that the useful evidence in 2WikiMultiHopQA may benefit from a broader selected memory set, while HotpotQA-Distractor is more sensitive to irrelevant selected context.

\begin{table}[t]
\centering
\small
\setlength{\tabcolsep}{4.5pt}
\renewcommand{\arraystretch}{1.05}
\begin{tabular}{@{}llccc@{}}
\toprule
Dataset & Setting & EM & F1 & ROUGE-L \\
\midrule
\multirow{5}{*}{HotpotQA-Dist.}
& \(K=2\)  & \textbf{38.42} & \textbf{50.93} & \textbf{50.82} \\
& \(K=4\)  & 38.34 & 50.83 & 50.71 \\
& \(K=6\)  & 36.69 & 49.20 & 49.10 \\
& \(K=8\)  & 35.15 & 47.47 & 47.36 \\
& \(K=10\) & 33.76 & 45.65 & 45.61 \\
\midrule
\multirow{5}{*}{2Wiki}
& \(K=2\)  & 49.40 & 55.69 & 55.66 \\
& \(K=4\)  & 49.90 & 56.41 & 56.29 \\
& \(K=6\)  & 50.90 & 57.21 & 57.08 \\
& \(K=8\)  & \textbf{51.10} & \textbf{57.42} & \textbf{57.32} \\
& \(K=10\) & 50.70 & 57.30 & 57.21 \\
\bottomrule
\end{tabular}
\caption{Effect of the selection budget \(K\) on HotpotQA-Distractor and
2WikiMultiHopQA. On HotpotQA-Distractor, increasing \(K\) beyond 2 reduces
answer quality, while on 2WikiMultiHopQA the score-based selection results
improve up to \(K=8\) and then saturate.}
\label{tab:k_sweep}
\end{table}

\section{Out-of-Distribution Transfer Test}
\label{app:ood_transfer}

We test whether the learned memory transfers beyond the QA adaptation distribution. Motivated by ICAE's two-stage setup~\cite{ge2024icae},
both SeDeM and a matched ICAE baseline are first compression-pretrained on SlimPajama and then adapted on the same multi-task QA mixture, with no QASPER examples used in either stage. QASPER differs from the adaptation data in domain (scientific papers vs.\ Wikipedia-style QA) and document length, making this an out-of-distribution transfer setting. Under
matched two-stage training, SeDeM reaches \(20.02\) token-F1 on
zero-shot QASPER, compared with \(11.30\) for ICAE. This suggests that decompressed hidden-state memories transfer better than
direct memory-token conditioning when the test domain and document structure
shift.

\section{Effect of LoRA on Decoder LLM}
\label{app:lora_ablation}

To isolate decoder adaptation, we compare a SeDeM-only model with no decoder LoRA against the same model with decoder LoRA in the selector-disabled setting. As shown in
Table~\ref{tab:lora_ablation}, the SeDeM-only model remains strong on both
datasets. These results indicate that the
method's gain is not merely a decoder-LoRA effect; LoRA provides an additional
adaptation benefit when useful, but the learned SeDeM itself carries the main signal.

\begin{table}[t]
\centering
\small
\setlength{\tabcolsep}{4pt}
\renewcommand{\arraystretch}{0.95}
\begin{tabular}{@{}lcc@{}}
\toprule
Dataset &
\shortstack{\\[-0.5ex]SeDeM\\+ frozen dec.} &
\shortstack{\\[-0.5ex]SeDeM\\+ adapted dec.} \\
\midrule
HotpotQA-Dist. & 62.55 & 69.78 \\
QASPER         & 23.36 & 24.37 \\
\bottomrule
\end{tabular}
\caption{
Decoder adaptation ablation in the selector-disabled setting. Both variants condition on all segment reconstructions, so these results are not efficiency-comparable to the top-\(K\) setting. We report F1 on HotpotQA-Distractor and QASPER.
}
\label{tab:lora_ablation}
\end{table}

\section{Granularity Ablation}
\label{app:codec_granularity_ablation}

We analyze how the granularity of the SeDeM affects reconstruction quality in the selector-disabled setting. Each document is divided into segments of length \(T\), and each segment is compressed into \(M\) memory slots with compression factor \(C\), such that \(T = M C\). This ablation varies two SeDeM knobs: the segment length \(T\), which controls how much local context is visible when a memory block is formed, and the compression factor \(C\), which controls how aggressively tokens are collapsed into memory slots.

\begin{figure}[t]
    \centering
    \includegraphics[width=\columnwidth]{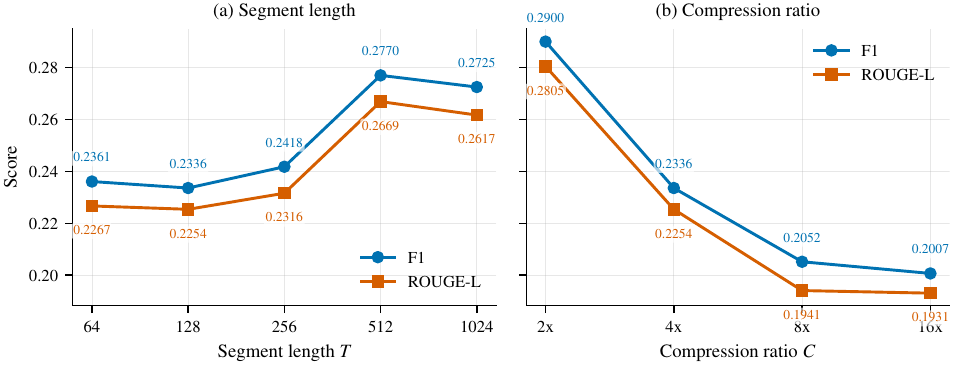}
    \caption{SeDeM granularity ablation on QASPER under selector-disabled evaluation. 
    Left: varying segment length \(T\) at fixed compression factor \(C=4\), which keeps the total memory budget fixed. 
    Right: varying compression factor \(C\) at fixed segment length \(T=128\), where larger \(C\) applies stronger compression and uses fewer memory slots. 
    The results show that moderate segment lengths improve SeDeM quality, while overly aggressive compression produces a monotonic quality decay.}
    \label{fig:codec_granularity_ablation}
\end{figure}

Figure~\ref{fig:codec_granularity_ablation} shows that these two knobs have different effects. Increasing the compression factor \(C\) leads to a monotonic quality decay, as fewer memory slots are available to preserve document information. This sweep should be interpreted as a cost--quality trade-off, since stronger compression also reduces the total memory budget. In contrast, the segment-length sweep keeps \(C=4\) fixed, so the total memory budget remains constant. Under this controlled setting, performance improves as \(T\) increases up to \(512\), suggesting that the SeDeM benefits from forming memory slots over a wider local context. The slight decline at \(T=1024\) indicates saturation: overly long segments may mix less related document regions into the same compression unit, reducing the specificity of the reconstructed states for question answering.

\section{Controlled Comparisons: Selection, Retrieval, and Compression}
\label{app:controls}

This appendix reports the controlled experiments that separate supervised
selection, compression--decompression quality, and raw-text conditioning. A
concise summary is given in Section~\ref{sec:controls_summary}. Each experiment
below states its backbone and supervision condition; results from different
conditions should not be read as head-to-head comparisons.

\paragraph{Identical-input control.}
We compare SeDeM and ICAE after giving both methods exactly the same
gold-selected input segments, using the same Llama-3.2-3B backbone, on the
complete HotpotQA-Distractor validation set. SeDeM obtains \(64.95\) F1,
compared with \(60.03\) for ICAE. Because the two methods receive an identical
input, this control isolates the compression and decoder-conditioning pathway
from input selection.

\paragraph{Full-bank (selector-disabled) controls.}
These controls disable the selector at evaluation and decompress the complete
memory bank, so they are quality-oriented controls rather than the selective
low-latency operating point. Both use Llama-3.2-1B. On HotpotQA-Distractor, a
Stage-2 variant trained \emph{without} evidence-based data filtering or an
evidence-ranking loss obtains \(59.17\) F1 / \(45.05\) EM on the full
validation set. On 2WikiMHQA, disabling selection at evaluation and
decompressing all blocks gives \(67.28\) F1 / \(61.36\) EM; this model's
Stage-2 training used evidence-related filtering and ranking supervision, so
its comparison against methods without equivalent supervision (e.g., ComprExIT)
is not supervision-matched. These full-bank results isolate the
compressor--decompressor pathway from selective retrieval at inference.

\paragraph{Distant-supervision control.}
An answer-string-based distant-supervision variant of the selector obtains
\(51.78\) F1 on HotpotQA-Distractor, within approximately one point of the
learned-\(K{=}2\) result of \(50.93\). This suggests that selector training
does not strictly require gold block-level evidence labels.

\paragraph{Raw-text RAG with matched retrieval supervision.}
We compare against raw-text retrieval baselines that use the same ranking
architecture and evidence supervision as SeDeM's selector.
On HotpotQA-Distractor with Llama-3.2-3B, a raw-text reader trained and
evaluated on the gold supporting paragraphs obtains \(84.21\) F1 / \(71.18\)
EM; we treat this as a lossless raw-text reference rather than a compression
baseline. A trained raw-text paragraph retriever with the same supervision
obtains \(66.31\) F1 / \(54.49\) EM at \(K{=}2\) and \(72.59\) F1 at
\(K{=}10\). The designated main 3B SeDeM configuration obtains \(58.30\) F1
(Table~\ref{tab:main_qa_results}), and under identical gold-selected input
SeDeM obtains \(64.95\) F1 versus \(60.03\) for ICAE.
On 2WikiMHQA with Llama-3.2-1B, the matched raw-text RAG baseline obtains
\(59.17\) F1 / \(53.51\) EM at \(K{=}2\) and \(46.74\) F1 / \(41.97\) EM at
\(K{=}10\), despite retrieval recall increasing to \(1.0\); the SeDeM
learned-\(K{=}2\) result is \(55.69\) F1. (The RAG \(K{=}2\) value of
\(59.17\) on 2WikiMHQA is numerically equal to, but distinct from, the
HotpotQA-Distractor full-bank control above.)
At $K=2$, raw-text RAG is stronger in answer quality in these comparisons. SeDeM's system-level distinction is that it stores reusable,
query-independent latent memories and does not reprocess selected raw text
from the decoder input layer for every query; we present this as a
quality--efficiency and system-design trade-off rather than superiority over
RAG.

\paragraph{Boundary-shift sensitivity.}
SeDeM does not require naturally occurring paragraphs, sections, or documents
as segments; it requires only bounded contiguous computational windows, and
arbitrary continuous text can be segmented this way. To quantify sensitivity
to window placement, we shift every segmentation boundary by 64 tokens while
preserving the number of windows. For the 1B HotpotQA-Distractor full-context
control, this changes F1 from \(59.17\) to \(57.87\), a reduction of \(1.30\)
points. Across five full-context configurations on HotpotQA-Distractor and
2WikiMHQA, the reduction ranges from \(0.81\) to \(2.14\) F1. We interpret
this as moderate sensitivity to window alignment rather than dependence on
naturally structured input. Dependencies spanning many distant, independently
compressed windows remain a limitation of independent segment compression.